\documentclass[runningheads]{llncs}

\usepackage{eccv}

\usepackage{eccvabbrv}

\usepackage{graphicx}
\usepackage{booktabs}

\usepackage[accsupp]{axessibility}  

\usepackage{hyperref}

\usepackage{orcidlink}

\usepackage{xparse}

\NewDocumentCommand{\citep}{O{} O{} m}{\cite{#3}}
\NewDocumentCommand{\citet}{O{} O{} m}{\cite{#3}}

\usepackage{tabularx}
\usepackage{array}
\usepackage{booktabs}
\usepackage{makecell}
\usepackage{algorithm}
\usepackage{algpseudocode}
\usepackage{wrapfig}

\usepackage{upgreek}

\renewcommand{\paragraph}[1]{{\vspace{1mm}\noindent \bf #1}}

\begin{document}

\title{Air Quality Downscaling \\ with Station-Guided Pseudo-Supervision}


\author{
Guorun Wang \inst{1} \and
Simone Foti \inst{1} \and
Andreas D. Demou \inst{2} \and
Leonidas Kotoulas \inst{1} \and
Theodoros Christoudias \inst{2} \and
Alexandros Koliousis \inst{3} \and
Mihalis Nicolaou \inst{4} \and
Stefanos Zafeiriou \inst{1} 
}

\authorrunning{G.~Wang et al.}

\institute{
Imperial College London, UK \and
The Cyprus Institute, Cyprus \and
Northeastern University London, UK \and
University of Cyprus, Cyprus
}

\maketitle

\begin{abstract}
Super-resolving coarse atmospheric fields to local PM$_{2.5}$ variations is uniquely challenged by a mismatch in spatial support: while pixels represent regional averages, ground-truth observations are discrete, unaligned samples of a continuous spatial signal. To bridge this gap, we present a station-guided framework for high-resolution PM$_{2.5}$ downscaling over Europe. Taking coarse CAMS atmospheric composition fields alongside heterogeneous side information (i.e., human activity, land cover, elevation, satellite aerosol observations, and wind fields) our framework jointly super-resolves ($\times 40$, $\approx$ 1~km) and bias-corrects CAMS rasters, without relying on temporal sequence modelling.
To address the challenge of densely supervising our multi-scale transformer network with sparse in-situ data, we introduce a time-agnostic propagation strategy that utilises spatial Gaussian blending of interpolated OpenAQ observations. Extensive qualitative and station-level evaluations across Europe demonstrate that our model recovers fine-grained spatial structures and effectively mitigates localised CAMS biases. 

\end{abstract}

\section{Introduction}
Climate change is reshaping atmospheric conditions in ways that directly influence air quality through complex interactions between meteorology, atmospheric chemistry, and human activities. As climatic conditions evolve, changes in temperature, circulation patterns, and the frequency of extreme events can affect the formation, transport, and dispersion of air pollutants, creating new challenges for air-quality management and forecasting. Air quality remains a major environmental and public health concern, affecting billions of people worldwide and contributing significantly to respiratory and cardiovascular disease burdens~\cite{goldsborough2023pollution}. Accurate and timely predictions are therefore essential for supporting public health interventions, informing environmental policy, and enhancing our understanding of Earth system processes.
In this context, machine-learning-based forecasting systems offer a promising pathway for improving the prediction of pollution episodes and informing public health decision-making.


The rapid development of machine learning (ML) systems for scientific prediction and decision-making has ushered in a new era for environmental forecasting~\citep{rumelhart1986learning, wolf-etal-2020-transformers, openai2023gpt}. In meteorology, this progress has been particularly evident in the emergence of advanced AI-based weather forecasting systems, which have achieved remarkable performance in predicting dynamical atmospheric variables~\citep{mccann1992neural, marzban1996neural, tangang1997forecasting, kuligowski1998experiments, deo1998real, hsieh1998applying, spellman1999application, kolehmainen2000forecasting, bi2023accurate, chen2023fuxi, lam2023learning, chen2025operational, han2024fengwu}. However, extending these successes from weather prediction to atmospheric-composition forecasting remains challenging. Unlike conventional weather forecasting, atmospheric-composition prediction must account not only for meteorological transport but also for emissions, chemical transformations, deposition, and human activities, making it substantially more complex and computationally demanding~\citep{bodnar2025foundation}. This challenge is especially important for air pollution, which directly affects human health and ecosystems~\citep{world2021global, cams2025report}. 

To tackle this, Aurora~\citep{bodnar2025foundation}, a foundation model for the Earth system, shows a pathway for moving beyond weather forecasting towards more general Earth-system prediction, including atmospheric composition.
ML has demonstrated strong capabilities in learning the complex nonlinear relationships between meteorology, emissions, and pollutant concentrations, an approach that has shown promising improvements in both predictive accuracy and computational efficiency, marking a paradigm shift from purely physics-based models toward ML-based frameworks~\citep{shi2021abrupt, teng202372, qiu2023regional, han2023capability, koo2024development, hossen2025ode, wang2025spatiotemporal, rautela2025unequal}.
Despite these promising developments, current methods such as Aurora typically operate at spatial resolutions of \(0.25^\circ\text{–}0.4^\circ\) (\(\approx 25\text{–}40\,\text{km}\)), which is still too coarse to capture pollution hotpspots and regional variations. Hence, an important direction is spatial downscaling (super-resolution), which aims to produce point-level predictions from coarse global reanalyses. 

On the data side, air quality monitoring and forecasting rely on two main data sources with complementary characteristics. On the one hand, forecast and analysis data such as those from the Copernicus Atmosphere Monitoring Service (CAMS) provide spatially complete and consistent estimates of atmospheric composition, particularly where observational data coverage is low or for atmospheric pollutants for which no direct observations are available~\citep{inness2019cams_reanalysis, cams_global_2021, cams_european_2021}. Reanalysis data combine physical models with past observations through a process called data assimilation. 
On the other hand, ground-based monitoring stations such as EBAS~\citep{torseth2012introduction} and OpenAQ~\citep{OpenAQ} provide highly accurate point-level measurements of pollutants like fine particulate matter (PM$_{2.5}$), regarded as the ground truth for model evaluation.
However, each data source has its own limitations.
For reanalysis and forecast products, the spatial resolution is relatively coarse (approximately 0.4$^\circ$, or about 40~km). Moreover, studies have reported that CAMS  exhibits systematic biases relative to ground-based measurements, often showing overestimation of particulate matter concentrations~\citep{eskes2024evaluation, gualtieri2025assessing}. 
In contrast, the station measurements are spatially sparse and unevenly distributed, with higher densities in particular areas and limited coverage in rural or remote areas. This spatial heterogeneity makes it difficult to construct high-resolution air-quality fields solely from station data.

The key challenge, therefore, lies in bridging these two data sources and combining the spatial completeness of CAMS with the precision of ground stations to produce high-resolution, bias-corrected air-quality estimates. Auxiliary variables that encode physical and environmental characteristics can provide additional information in this process. Incorporating high-resolution side information enables the model to learn physically-grounded relationships, and to propagate point observations from stations into the coarse CAMS grids, thereby improving downscaling performance. 

In this study, we focus on PM$_{2.5}$ downscaling over Europe using OpenAQ station measurements as the ground-truth reference. Starting from the 0.4$^\circ$ CAMS forecast fields as the prior, we integrate multiple auxiliary information sources: human activity, land use, elevation, aerosol satellite observations, and wind to guide the learning process. The objective is to construct a model that effectively propagates the accuracy of the OpenAQ dataset to the broad spatial coverage of CAMS, achieving physically consistent and spatially detailed PM$_{2.5}$ estimates, thereby enabling a resolution enhancement from 0.4$^\circ$ to 0.01$^\circ$. 

Our \textbf{main contributions} are summarized as follows:
\begin{itemize}
    \item \textbf{Task formulation.} We recast fine-grained PM\textsubscript{2.5} estimation as a \emph{time-agnostic, joint downscaling and bias-correction} problem rather than temporal forecasting. Given coarse CAMS fields and time-matched side information, our model produces $\times 40$ 
    (${\approx}1$~km) 
    bias-corrected PM\textsubscript{2.5} maps at any single timestamp, without temporal sequence modeling, allowing it to generalize to unseen times and locations.

    \item \textbf{Gaussian-kernel pseudo-label propagation.} To densely supervise the network from spatially sparse, grid-unaligned station data, we introduce a propagation scheme that spreads OpenAQ observations into a dense target and blends it with the CAMS prior where station support is weak. This turns a few thousand point measurements into pixel-wise supervision representing the ground-truth accuracy directly into the coarse field.

    \item \textbf{Continental-scale framework and evidence.} We propose STARQ (STation AiR Quality), our continental-scale framework over all Europe, fusing eight heterogeneous geospatial and atmospheric sources within a multi-scale SegFormer backbone. Under a strict spatiotemporal split (unseen timestamps \emph{and} unseen stations), it consistently outperforms both CAMS and S-MESH$^*$, our implementation of S-MESH~\cite{shetty2025daily}, the XGBoost-based state-of-the art method for station-guided downscaling. 
\end{itemize}




\begin{figure*}[t]
    \centering
    \includegraphics[width=\linewidth]{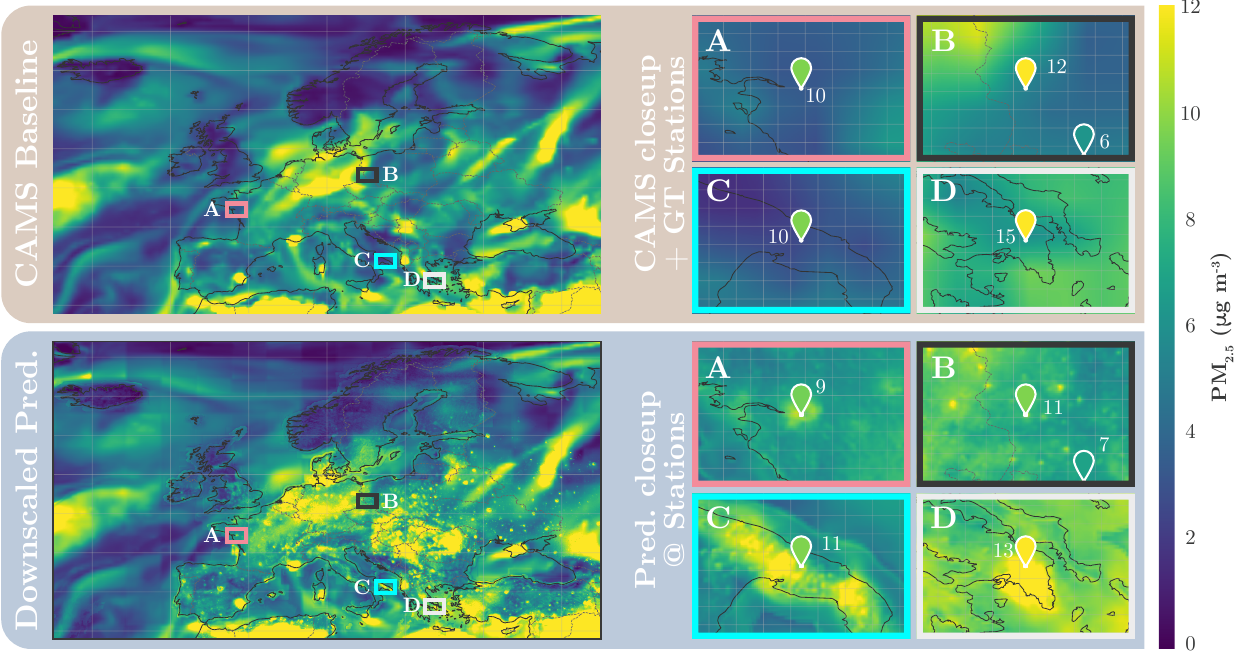}
    \caption{PM$_{2.5}$ over Europe from CAMS forecast (\emph{top}) and downscaled by STARQ (\emph{bottom}). Close-ups (\emph{right}) are reported alongside ground truth station values and our corresponding predictions.}
    \label{fig:eu_pm25_zoomin}
\end{figure*}


\section{Related Work}

\paragraph{Downscaling for PM\textsubscript{2.5}.}
High spatiotemporal-resolution PM\textsubscript{2.5} concentration data are essential for air-pollution assessment and research. Over the past decade, PM\textsubscript{2.5} downscaling methods have evolved from traditional statistical approaches to modern ML and deep learning frameworks. Early PM\textsubscript{2.5} downscaling methods typically use coarse-scale model outputs as inputs and transform them into high-resolution estimates through statistical relationships. For example, \citet{varga2020time} downscale CAMS PM\textsubscript{2.5} using CAMS as a prior together with urban monitoring data from Budapest. However, the inherent linearity of such approaches may introduce systematic biases. In contrast, geostatistical methods such as kriging \citep{oliver1990kriging} explicitly model spatial autocorrelation, although they require dense monitoring networks to perform reliably.

Some methods also address missing auxiliary data. MODIS aerosol optical depth (AOD), a commonly used source of side information, often contains missing values. To address this issue, \citet{lv2017daily} employ a Bayesian-based statistical downscaler to model the spatio-temporal linear relationships between AOD and PM\textsubscript{2.5} and to fill missing AOD values. In parallel, \citet{yang2020mapping} downscale the 0.1° ACAG PM\textsubscript{2.5} product to approximately 300 m using a cascade random forest model driven by elevation, land-cover data, and AOD. \citet{pu2021ground} impute 1 km MAIAC AOD using a quantile regression forest and then use a gradient boosting machine algorithm to estimate ground-level PM\textsubscript{2.5}. Overall, these methods improve resolution and accuracy to some extent, but they typically rely on extensive feature engineering and strong modelling assumptions.

Recent studies further explore PM\textsubscript{2.5} downscaling and forecasting across different regions and temporal scales. \citet{shi2021abrupt} analyze hourly air-pollutant data across 11 global cities during the COVID-19 lockdowns, using station-level spatial resolution and hourly temporal resolution. \citet{teng202372} develop a hybrid GNN--LSTM model for 72-hour PM\textsubscript{2.5} forecasting over the Beijing--Tianjin--Hebei region, using hourly data from 2016--2020 at station-level spatial resolution. In the same region, \citet{qiu2023regional} use data from 2020--2022 at 3-hour temporal and 9 km spatial resolution for PM\textsubscript{2.5} forecasting. \citet{koo2024development} focus on PM\textsubscript{2.5} forecasting in Seoul, South Korea, operating at 6-hour temporal and 27 km spatial resolution. \citet{hossen2025ode} conduct hourly PM\textsubscript{2.5} forecasting across 18 stations in the Taipei metropolitan area, using data from 2014--2020 at 1-hour temporal resolution. \citet{rautela2025unequal} estimate global daily PM\textsubscript{2.5} concentrations at 0.5° × 0.625° resolution from 1980--2023, achieving high accuracy and revealing persistent pollution extremes in South and East Asia. However, these studies either focus on relatively small regions or produce results at comparatively coarse spatial resolution. Many also depend on temporal information or forecasting-specific settings. These limitations motivate our time-independent downscaling task over a large European domain.

Recent European-scale downscaling methods further incorporate station observations. \citet{shetty2025daily} employs station-supervised XGBoost \citep{chen2016xgboost} to downscale CAMS regional PM\textsubscript{2.5} forecasts into 1 km daily PM\textsubscript{2.5} maps, while \citet{guion2026high} trains a framework to generate 500 m annual air-quality maps for multiple pollutants across Europe. However, neither method leverages hourly OpenAQ station observations for timestamp-level supervision, which would enable PM\textsubscript{2.5} downscaling at arbitrary hours. In contrast, our work addresses this gap by incorporating hourly station observations to support timestamp-level PM\textsubscript{2.5} downscaling.

\paragraph{Convolutional and Transformer Architectures.}
Neural networks can automatically extract complex spatial feature relationships and demonstrate strong nonlinear fitting capabilities in downscaling tasks. For PM\textsubscript{2.5}, \citet{xue2019spatiotemporal} successfully applied spatial convolution to the downscaling of meteorological variables, demonstrating that spatial convolution is suitable for refining continuous fields such as air pollution. Convolutional architectures are powerful tools for image processing and have been widely applied to super-resolution and climate data downscaling \citep{wang2021deep, bano2020configuration, geiss2022downscaling, xiao2022generating, ashiotis2022ai}.

Encoder-decoder networks, such as U-Net and Transformer-based models, can effectively fuse multi-scale features \citep{ronneberger2015u,lee2025conditional, vaswani2017attention, xie2021segformer}. We recognize the strong performance of convolutional and transformer architectures in downscaling tasks, and our approach builds on their respective advantages. However, existing studies often fail to achieve sufficiently fine spatial resolution, such as around 1 km, are limited in their ability to cover large regions, such as intercontinental or global domains, or rely on temporal dependencies that require time-series contextual information. In contrast, our method achieves a very high spatial resolution of 0.01\textdegree{} while maintaining broad spatial coverage across the whole of Europe, and it does not depend on temporal inputs. This means that, given the required spatial information at any point in time, our model can perform downscaling accordingly.

\paragraph{Gaussian Kernel Interpolation for PM\textsubscript{2.5}.}
 The Gaussian downscaling method, originally introduced by \citet{shen2014global, sorensen1998sensitivity} to characterize the spatial dispersion of pollutants, was adopted by \citet{shen2017urbanization} to refine coarse 50 km WRF/Chem PM\textsubscript{2.5} outputs to a 5 km resolution. Thus, the importance of Gaussian-related formulas for climate studies is evident. Moreover, the reliability of interpolation methods has already been demonstrated in reconstructing temperature fields \citep{liu2018new}. In this study, we used the Gaussian kernel to perform interpolation on the propagation of station measurements.

\section{Data}

\paragraph{Data Sources.}
\label{sec:data_sources}
We use eight geospatial and atmospheric raster
datasets as inputs. Specifically, CAMS global
atmospheric composition forecasts \citep{cams_global_2021} serve as the low-resolution
PM$_{2.5}$ rasters to downscale and bias-correct.
Global Human Settlement Layer (GHSL) provides anthropogenic activity information such as the built-up surface
(GHS-BUILT-S) \citep{GHS_BUILT_S_R2023A}, built-up volume (GHS-BUILT-V) \citep{GHS_BUILT_V_R2023A}, and population
(GHS-POP) \citep{GHS_POP_R2023A}. In addition, CORINE Land Cover (CLC) \citep{CLMS_CLC2018_raster100m}characterises
the land-use conditions, EU-DEM \citep{EUDEM_Eurostat_2016} captures the topography, MODIS MCD19A2
Aerosol Optical Depth (AOD) \citep{MCD19A2_V061_LPDAAC} measures the satellite-observed aerosol
loading, and ERA5 10-m \citep{ERA5_hourly_CDS_2023} provides the wind components for inferring atmospheric
transport and dispersion.
The spatial and temporal resolutions of these datasets are summarised
in Table~\ref{tab:dataset_info}. All input variables are spatially
aligned and resampled onto a common $0.01^\circ$ European grid, while
time-varying variables are matched to the corresponding OpenAQ
observation timestamps. Detailed descriptions of the datasets and
their preprocessing procedures are provided in Sections A and B of the Supplementary
Material.

While all the aforementioned input datasets are regular raster data, the ground-truth PM$_{2.5}$ measurements used for supervision are sparse point observations obtained from OpenAQ, an open-source global air-quality data aggregation and sharing platform. The station measurements in OpenAQ come from various providers, including governments, research institutions, and open community networks. For the purpose of this study we select stations within Europe, regardless of their provider. In addition, we curate a significantly smaller and computationally tractable dataset from 2018, 2020, and 2023, covering the Italian peninsula and part of the Balkans, (spanning $36^\circ\mathrm{N}$--$47^\circ\mathrm{N}$ and $6^\circ\mathrm{E}$--$19^\circ\mathrm{E}$). We hereafter refer to this subset as the Italian dataset.

\begin{table}[t]
\centering
\scriptsize
\setlength{\tabcolsep}{3pt}
\renewcommand{\arraystretch}{1.0}

\begin{tabularx}{\linewidth}{@{}
    >{\raggedright\arraybackslash}X
    >{\raggedright\arraybackslash}p{2.8cm}
    >{\raggedright\arraybackslash}p{2.9cm}
@{}}
\toprule
\textbf{Dataset} &
\textbf{\makecell[l]{Spatial\\Resolution}} &
\textbf{\makecell[l]{Temporal\\Resolution}} \\
\midrule

CAMS forecast
& $0.4^\circ$
& hourly \\

GHSL Built-up Surface (GHS-BUILT-S)
& 3 arcsec
& 2015, 2020, 2025 \\

GHSL Built-up Volume (GHS-BUILT-V)
& 3 arcsec
& 2015, 2020, 2025 \\

GHSL Population Grid (GHS-POP)
& 3 arcsec
& 2015, 2020, 2025 \\

Grouping Land Use CLC
& 100 m
& 2012, 2018 \\

EU-DEM
& 1 arcsec
& static \\

MODIS MCD19A2
& 1 km
& daily \\

ERA5 wind (single levels)
& $0.25^\circ$
& hourly \\

OpenAQ
& \mbox{point measurements}
& provider-dependent \\

\bottomrule
\end{tabularx}

\caption{Spatial and temporal resolutions of the datasets.}
\label{tab:dataset_info}
\end{table}

\begin{figure}[t]
    \centering

    \begin{subfigure}[t]{0.57\textwidth}
        \centering
        \includegraphics[width=\linewidth]
        {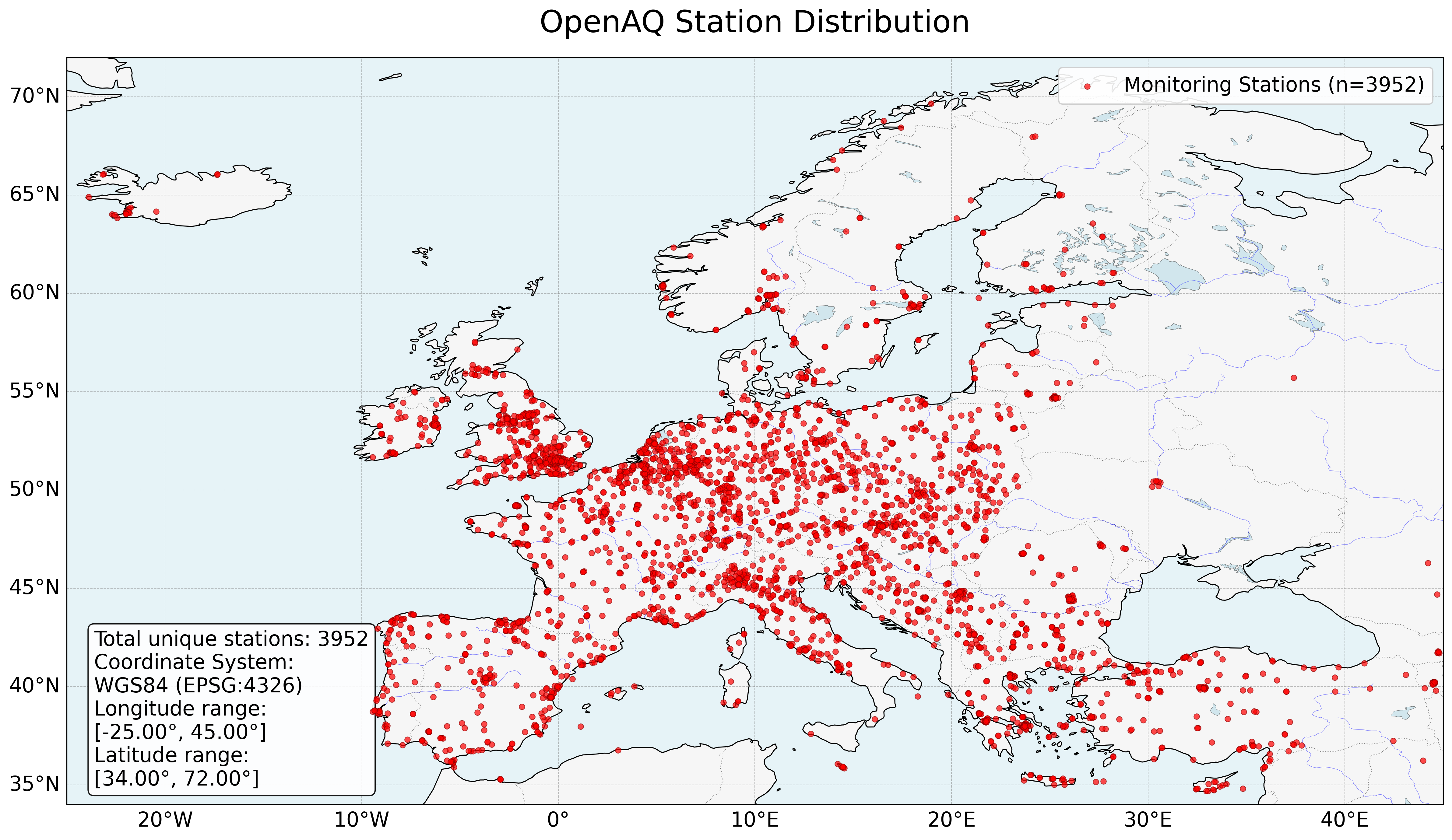}
        \caption{Spatial distribution of OpenAQ monitoring stations
        over the European domain.}
        \label{fig:openaq_station_distribution}
    \end{subfigure}
    \hfill
    \begin{subfigure}[t]{0.41\textwidth}
        \centering
        \includegraphics[width=\linewidth]
        {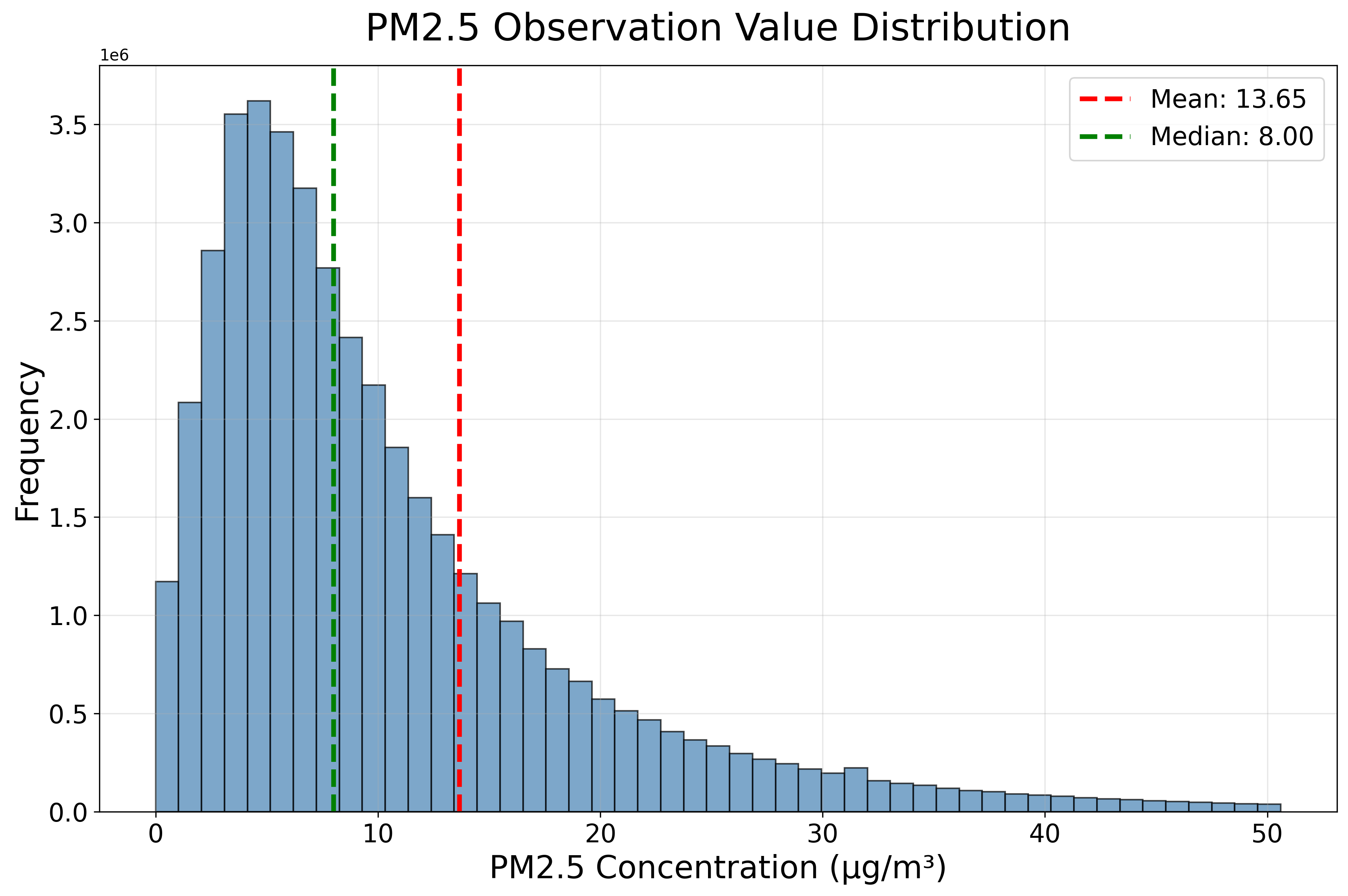}
        \caption{Distribution of OpenAQ PM$_{2.5}$ observations.}
        \label{fig:openaq_value_distribution}
    \end{subfigure}

    \caption{Spatial coverage and value distribution of the OpenAQ
    PM$_{2.5}$ observations used in this study.}
    \label{fig:openaq_data_statistics}
\end{figure}

Since our method produces a super-resolved PM$_{2.5}$ rasters with a 0.01$^\circ$ spatial resolution, we project station measurements to this resolution by averaging temporally concurrent observations from all stations within the same pixel. 
Following this projection, the dataset contains
3,952 unique grid-aligned stations over Europe. 
As can be observed in \cref{fig:openaq_station_distribution,fig:openaq_value_distribution}, not only is the stations' spatial distribution highly non uniform, but the PM$_{2.5}$ observations are also strongly right-skewed, with a
median of 8.00~$\upmu\mathrm{g}/\mathrm{m}^3$ and a mean of
13.65~$\upmu\mathrm{g}/\mathrm{m}^3$. Most observations lie in low-to-moderate
concentration ranges, while a small number of high-concentration and extreme
values form a long tail. These properties make the learning problem
challenging: the model must generalize across uneven station coverage while
remaining robust to rare high-pollution events.

\section{Methodology}

\subsection{Model Architecture}
Since the input is a multi-channel geospatial raster, we first apply a shallow convolutional network to project the input channels into a fixed feature dimension. Then, the projected feature map is passed to a four-stage hierarchical Transformer based on the SegFormer encoder. 
The multi-scale features computed by the four encoder stages are projected to a common decoder dimension, upsampled to the same spatial resolution, concatenated, and fused. The fused representation is then upsampled to the original patch resolution and passed through a convolutional regression head to predict a single-channel PM$_{2.5}$ concentration map.

\subsection{Gaussian-Kernel Pseudo-Label Propagation}
\label{sec:pseudo_label}

Sparse station observations provide accurate PM$_{2.5}$ measurements only at a limited number of monitoring sites, whereas our goal is to produce dense pixel-wise predictions. To bridge this gap, we introduce a station-guided pseudo-label propagation strategy. The method uses a Gaussian kernel to propagate point-level station observations to nearby pixels and blends the resulting interpolated field with a coarse baseline estimate in regions with station support.

Let $\Omega$ denote the spatial domain and let $\mathcal{S}=\{\mathbf{s}_i\}_{i=1}^{N}$ be the set of monitoring stations with valid PM$_{2.5}$ observations $y_i$. For each pixel location $\mathbf{p}\in\Omega$, we denote the model prediction by $\hat{y}(\mathbf{p})$ and the baseline estimate, e.g., CAMS, by $b(\mathbf{p})$.

\paragraph{Gaussian station influence.}
Each station contributes to nearby pixels through a Gaussian influence function:
\begin{equation}
    w_i(\mathbf{p}) =
    \exp\left(
    -\frac{\|\mathbf{p}-\mathbf{s}_i\|_2^2}{2\sigma^2}
    \right),
    \label{eq:gaussian_weight}
\end{equation}
where $\sigma$ controls the spatial decay of station influence. The station-interpolated field is then computed as:
\begin{equation}
    \tilde{y}(\mathbf{p}) =
    \frac{
    \sum_{i=1}^{N} w_i(\mathbf{p}) y_i
    }{
    \sum_{i=1}^{N} w_i(\mathbf{p}) + \epsilon
    },
    \label{eq:station_interpolation}
\end{equation}
where $\epsilon$ is a small constant for numerical stability.

\paragraph{Pseudo-label generation.}
We further define a cumulative station-confidence map:
\begin{equation}
    W(\mathbf{p}) =
    \max
    \left(0, \min \left(1, 
    \sum_{i=1}^{N} w_i(\mathbf{p}) \right)
    \right),
    \label{eq:station_confidence}
\end{equation}
which measures how strongly a pixel is supported by nearby stations. The final pseudo-label is obtained by blending the station-interpolated field with the baseline estimate:
\begin{equation}
    y_{\mathrm{pseudo}}(\mathbf{p}) =
    W(\mathbf{p})\tilde{y}(\mathbf{p})
    +
    \left(1-W(\mathbf{p})\right)b(\mathbf{p}).
    \label{eq:pseudo_label}
\end{equation}
At station locations, the pseudo-label is replaced by the observed ground truth, and these locations do not contribute to the pseudo-label supervision loss:
\begin{equation}
    y_{\mathrm{pseudo}}(\mathbf{s}_i) = y_i.
    \label{eq:station_override}
\end{equation}

\paragraph{Training objective.}
The training loss combines station-level supervision from observed measurements
with dense pseudo-label supervision over the remaining spatial domain:
\begin{align}
    \mathcal{L}_{\mathrm{station}}
    &=
    \frac{1}{N}
    \sum_{i=1}^{N}
    \left(
    \hat{y}(\mathbf{s}_i) - y_i
    \right)^2,
    \label{eq:station_loss}
    \\
    \mathcal{L}_{\mathrm{pseudo}}
    &=
    \frac{1}{|\Omega\setminus\mathcal{S}|}
    \sum_{\mathbf{p}\in\Omega\setminus\mathcal{S}}
    \left(
    \hat{y}(\mathbf{p}) -
    y_{\mathrm{pseudo}}(\mathbf{p})
    \right)^2,
    \label{eq:pseudo_loss}
    \\
    \mathcal{L}_{\mathrm{total}}
    &=
    \lambda_s \mathcal{L}_{\mathrm{station}}
    +
    \lambda_p \mathcal{L}_{\mathrm{pseudo}},
    \label{eq:total_loss}
\end{align}
where $\lambda_s$ and $\lambda_p$ control the relative contributions of
the station-level supervision and the dense pseudo-label supervision,
respectively.

\begin{figure}[!t]
    \centering

    \includegraphics[width=0.96\linewidth]{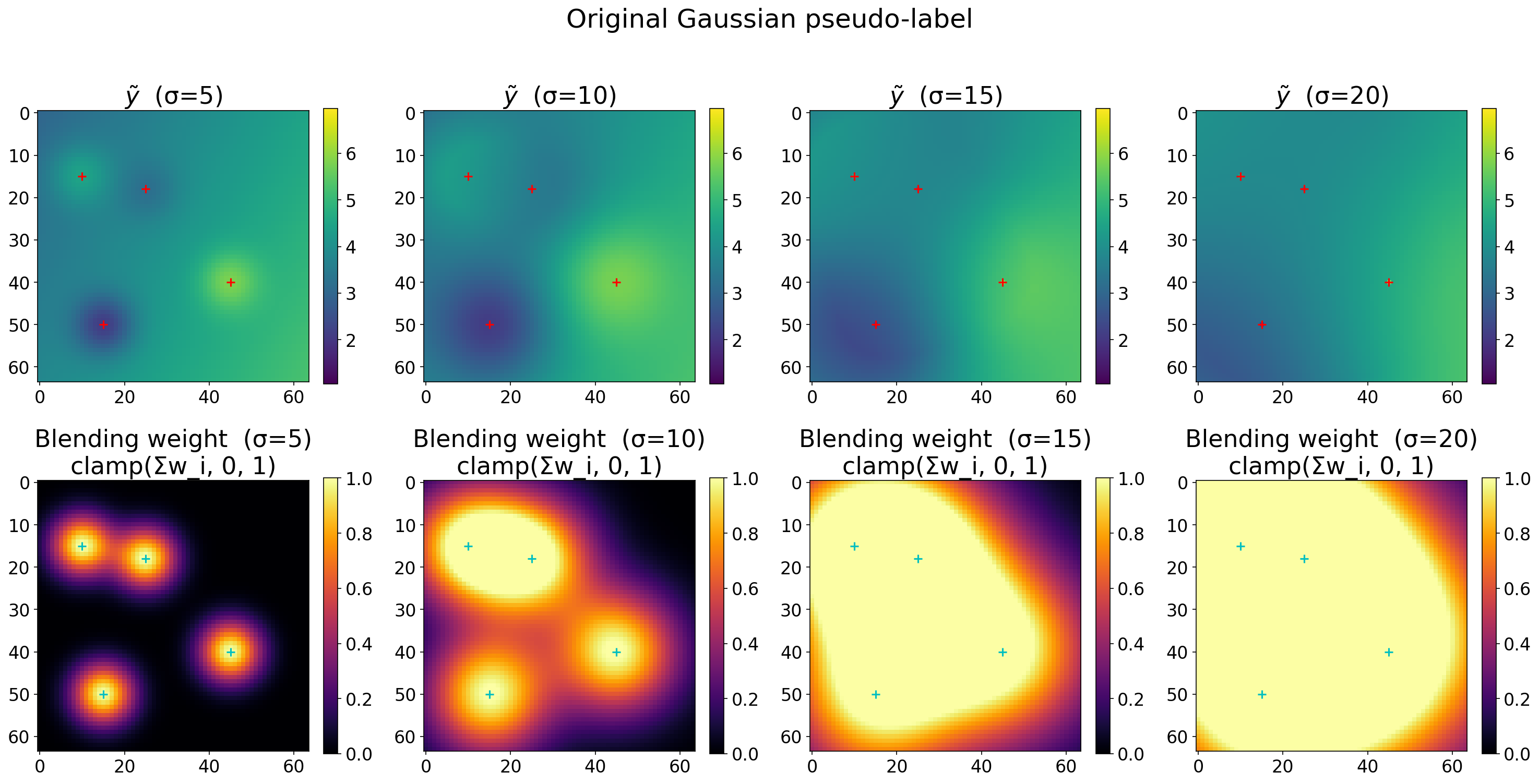}

    \smallskip
    \centerline{\small (a) Effect of kernel width $\sigma$.}

    \medskip

    \includegraphics[width=0.90\linewidth]{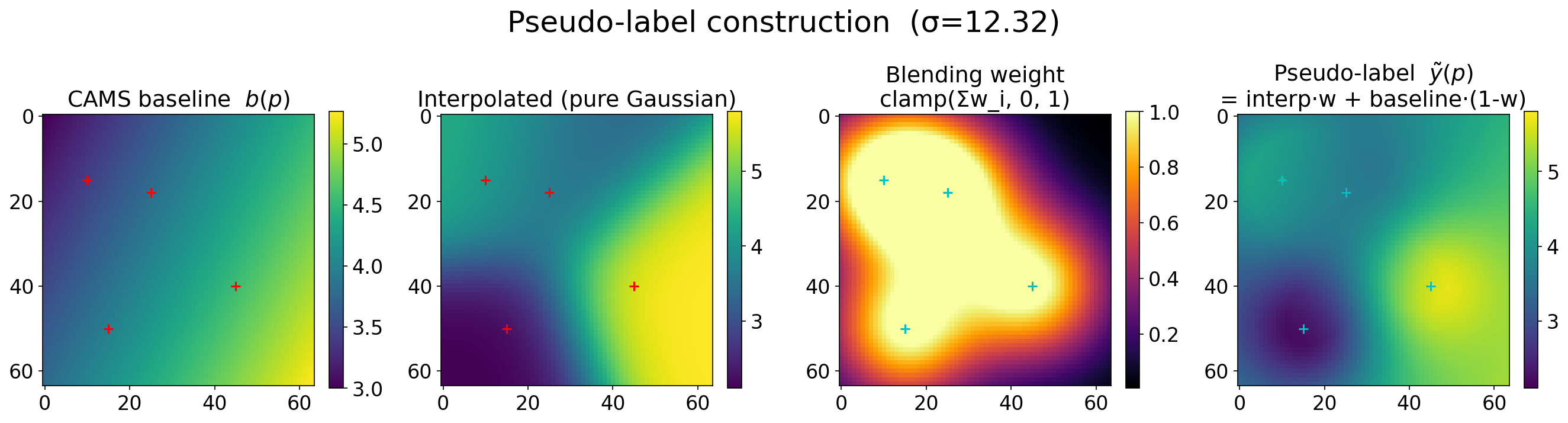}

    \smallskip
    \centerline{\small (b) Pseudo-label and loss-weight construction for the best hyperparameters.}

    \medskip

    \includegraphics[width=0.78\linewidth]{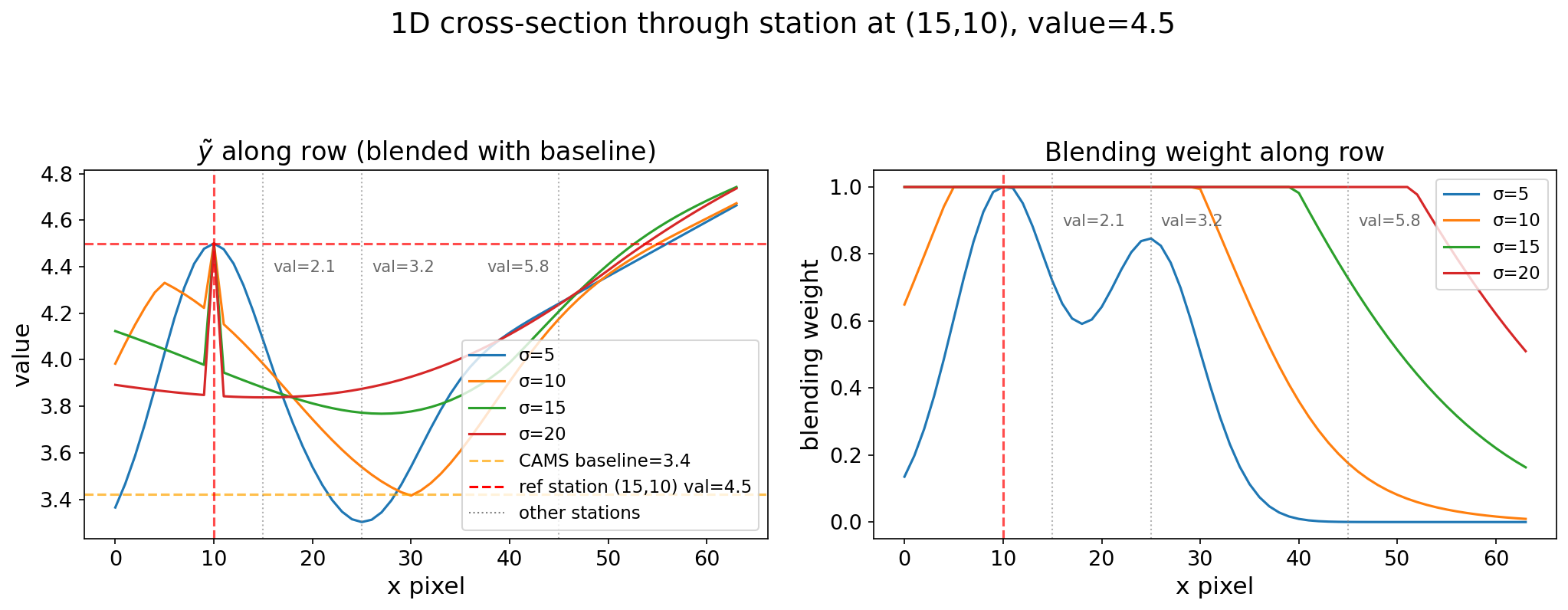}

    \smallskip
    \centerline{\small (c) One-dimensional cross-section.}

    \caption{
    Visualization of Gaussian pseudo-label propagation. Increasing $\sigma$
    expands station influence, while the pseudo-label blends station
    interpolation with the CAMS baseline.
    }
    \label{fig:kernel_visualization}
\end{figure}

We visualize the behaviour of the Gaussian pseudo-label propagation in
\cref{fig:kernel_visualization}. 
The one-dimensional cross-section further illustrates how the pseudo-label
transitions from station observations back to the CAMS baseline.

\section{Experiments}

\subsection{Implementation Details}
    
    \paragraph{Architecture and hyperparameters.}
        The encoder adopts a custom four-stage SegFormer configuration with hidden dimensions of [256, 512, 1280, 2048] and transformer depths of [3, 8, 27, 3]. Because the input channels correspond to meteorological, chemical, and geographical variables rather than natural RGB images, the model is trained from scratch by minimising \cref{eq:total_loss}. The best performing model is selected based on the epoch with the lowest validation MAE with respect to station data.
        
        Hyperparameter tuning was performed on the Italian dataset using Ray with Optuna, and using the validation MAE as optimization objective. The best configuration, subsequently used also on the full European dataset, consisted of a background sampling ratio of $0.176$, a learning rate of $1.31 \times 10^{-5}$, a weight decay of $1.95 \times 10^{-5}$, a Gaussian standard deviation of $\sigma=12.32$ pixels, as well as station- and pseudo-label-supervision weights of $\lambda_s=0.173$ and $\lambda_p=0.970$, respectively.

    \paragraph{Data splits and patch extraction.}
        To rigorously evaluate generalization, we perform a spatiotemporal split. Timestamps are split into train, validation, and test sets with a ratio of 70:20:10, while station locations are split with a ratio of 80:10:10. This ensures the model is evaluated on unseen side information and station locations. We extract $64 \times 64$ input patches (covering approximately $64 \times 64$\,km) cropped around monitoring stations with random spatial offsets. We also sample background patches without station supervision to encourage the model to preserve the large-scale spatial priors learned from the CAMS baseline. While during training patches are not necessarily centred with respect to a station, during validation and testing, the target station is always placed at the centre of the patch, and metrics are computed only at the site location. 

    \paragraph{Bucketed timestamp sampling.}
        Fully random patch sampling across all timestamps creates a severe I/O bottleneck, while sampling from a single timestamp destroys the temporal diversity required for stable gradients. We balance data locality and batch diversity using a bucketed timestamp sampling strategy ($K=4$). Timestamps are randomly shuffled and grouped into buckets of four. All patches associated with these timestamps are pooled, randomly shuffled, and sequentially yielded as mini-batches. This restricts data access to a small, cached subset of files while retaining temporal variation within each batch.

    \paragraph{Hardware.}
        All experiments were run on 4 NVIDIA H200 NVL GPUs. For the main experiments, we used a per-GPU batch size of \texttt{BATCH\_SIZE}=384. For hyperparameter tuning, we launched 20 concurrent Optuna trials on a single NVIDIA H200 NVL GPU.

\subsection{EU Downscaling Results}
The evaluation results for the EU dataset shown in Table \ref{tab:test_set_comparison} are measured in terms of mean absolute error (MAE), root mean squared error (RMSE), and coefficient of determination ($R^2$) between predicted values at station locations and corresponding ground-truth observations. Our model outperforms both the baseline CAMS Forecast and S-MESH$^*$, our implementation of S-MESH trained on the same data used to train STARQ.
We exhibit lower MAE and RMSE, as well as higher R\textsuperscript{2}. Following the methodology adopted in \citet{ashiotis2022ai}, we divided the CLC dataset into 15 classes for model training and evaluation. To more clearly demonstrate the advantages of our model, we further grouped these classes into five broader categories and evaluated the improvements achieved across heterogeneous land-use conditions.
As shown in Table~\ref{tab:landuse_broad_comparison}, STARQ consistently achieves the best performance across all broad land-use categories in terms of MAE, RMSE, and $R^2$. The improvement is observed not only in the dominant urban category, which contains more than half of the test observations, but also across industrial, agricultural, natural, and water-covered regions. The largest improvement over CAMS Forecast is observed in the water category, where STARQ reduces MAE from 6.611 to 4.986 and improves $R^2$ from $-0.392$ to 0.219. S-MESH$^*$ generally improves upon CAMS Forecast, but its gains are less consistent and substantially smaller than those achieved by STARQ. These results indicate that explicitly modelling spatial context provides more robust improvements across heterogeneous land-use conditions than point-wise nonlinear regression.

\begin{table}[t]
\caption{Performance comparison on the EU test set between the CAMS baseline, S-MESH$^*$, and our Gaussian-kernel pseudo-label propagation method.}
\label{tab:test_set_comparison}
\centering
\begin{tabular}{lccc}
\toprule
  & MAE $\downarrow$ & RMSE $\downarrow$ & R$^2$ $\uparrow$ \\
\midrule
CAMS Forecast       & 6.562 & 13.723 & 0.029 \\
S-MESH$^*$        & 6.485 & 13.080 & 0.118 \\
STARQ (ours)        & \textbf{5.873} & \textbf{12.126} & \textbf{0.242} \\
\bottomrule
\end{tabular}
\end{table}

\begin{table*}[t]
    \centering
    \caption{
    Performance comparison across broad land-use categories.
    MAE and RMSE are reported in $\upmu\mathrm{g}/\mathrm{m}^{3}$.
    Lower MAE and RMSE are better, while higher $R^2$ is better.
    The best result for each metric within each land-use category is
    highlighted in bold.
    }
    \label{tab:landuse_broad_comparison}

    \resizebox{\textwidth}{!}{
    \begin{tabular}{l r rrr rrr rrr}
        \toprule
        & &
        \multicolumn{3}{c}{\textbf{STARQ (ours)}} &
        \multicolumn{3}{c}{\textbf{S-MESH$^*$}} &
        \multicolumn{3}{c}{\textbf{CAMS Forecast}} \\
        \cmidrule(lr){3-5}
        \cmidrule(lr){6-8}
        \cmidrule(lr){9-11}

        \textbf{Land use}
        & N
        & MAE $\downarrow$
        & RMSE $\downarrow$
        & $R^2$ $\uparrow$
        & MAE $\downarrow$
        & RMSE $\downarrow$
        & $R^2$ $\uparrow$
        & MAE $\downarrow$
        & RMSE $\downarrow$
        & $R^2$ $\uparrow$ \\
        \midrule

        Urban
        & 199,020
        & \textbf{6.297}
        & \textbf{14.274}
        & \textbf{0.210}
        & 6.893
        & 15.213
        & 0.103
        & 6.945
        & 15.862
        & 0.025 \\

        Industrial
        & 70,342
        & \textbf{5.250}
        & \textbf{9.275}
        & \textbf{0.320}
        & 5.929
        & 10.288
        & 0.163
        & 6.005
        & 10.946
        & 0.053 \\

        Agriculture
        & 62,444
        & \textbf{5.431}
        & \textbf{8.797}
        & \textbf{0.254}
        & 5.891
        & 9.391
        & 0.150
        & 5.740
        & 9.787
        & 0.076 \\

        Natural
        & 28,506
        & \textbf{5.565}
        & \textbf{9.008}
        & \textbf{0.259}
        & 6.348
        & 10.476
        & $-0.002$
        & 6.348
        & 10.395
        & 0.013 \\

        Water
        & 16,621
        & \textbf{4.986}
        & \textbf{7.872}
        & \textbf{0.219}
        & 5.538
        & 8.602
        & 0.067
        & 6.611
        & 10.506
        & $-0.392$ \\

        No Data/Unknown
        & 1,105
        & \textbf{15.222}
        & \textbf{27.055}
        & \textbf{0.366}
        & 19.711
        & 34.391
        & $-0.024$
        & 24.535
        & 40.260
        & $-0.404$ \\

        \bottomrule
    \end{tabular}
    }
\end{table*}

For full-domain visualization, we apply overlapping sliding-window
inference with window size $W_s$ and overlap $O$. Predictions from
overlapping windows are merged using normalized weighted averaging,
where pixels near patch boundaries receive lower weights than those
near the patch centre. This boundary-aware blending suppresses patch
artifacts and produces a spatially smooth prediction over the entire
European domain. Further details of the padding, weighting, and
aggregation procedures are provided in Section C of the Supplementary Material.

\begin{figure*}[t]
    \centering
    \begin{tabular}{cc}
        \includegraphics[width=0.48\textwidth]{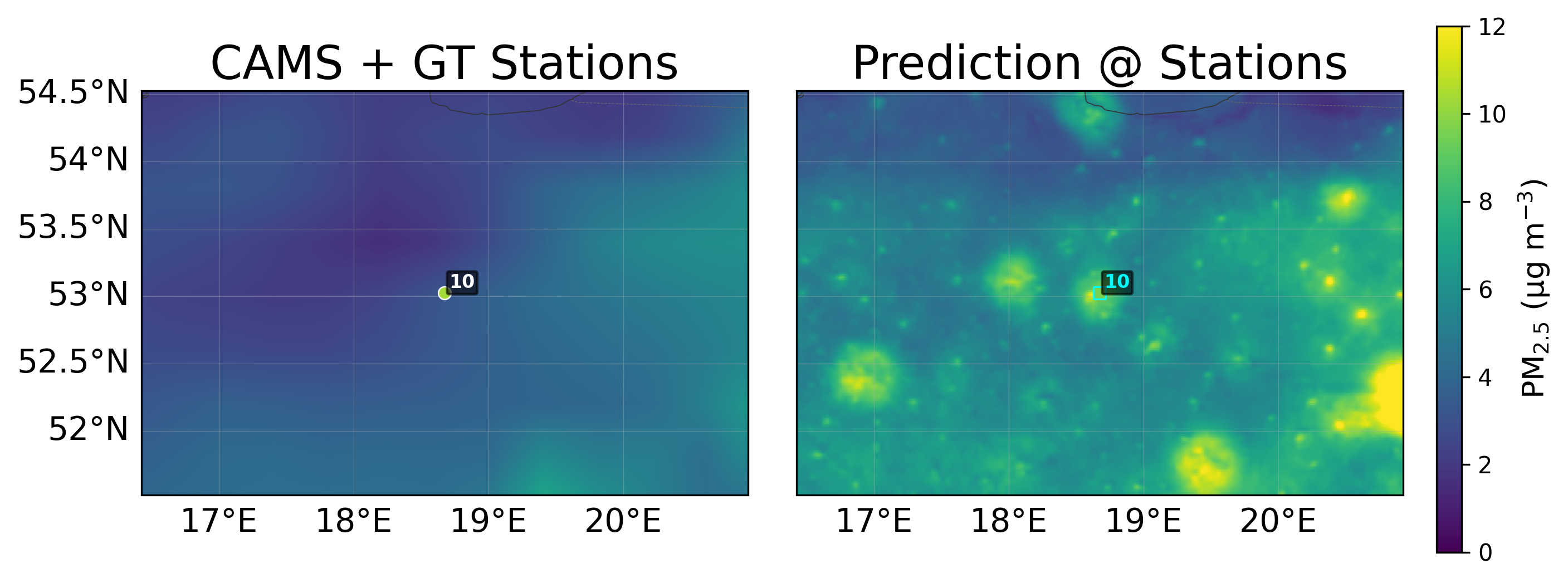} &
        \includegraphics[width=0.48\textwidth]{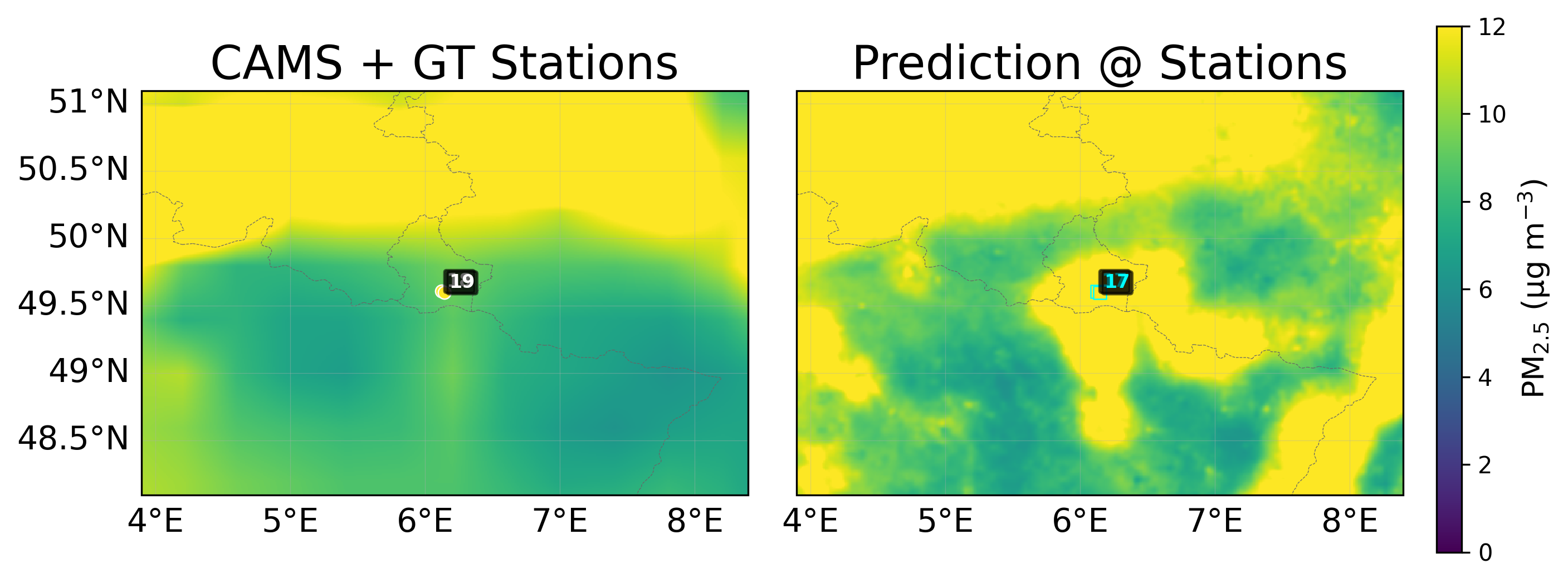} \\[2mm]
        \includegraphics[width=0.48\textwidth]{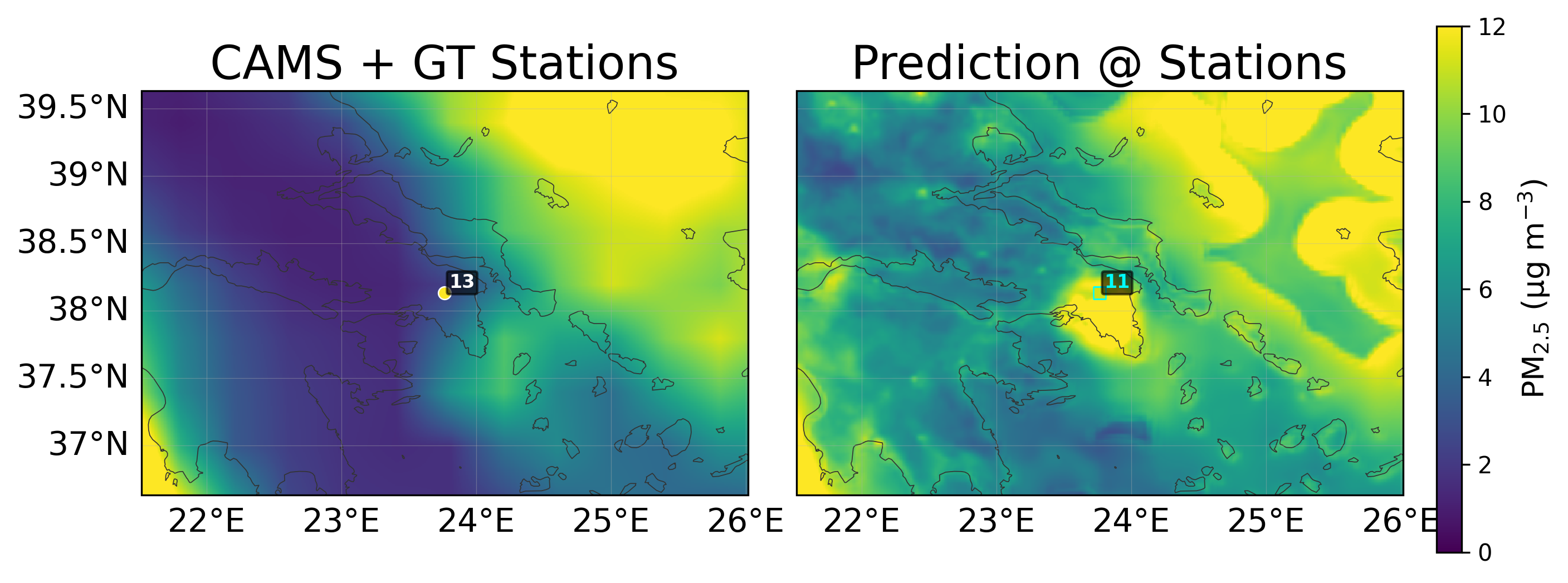} &
        \includegraphics[width=0.48\textwidth]{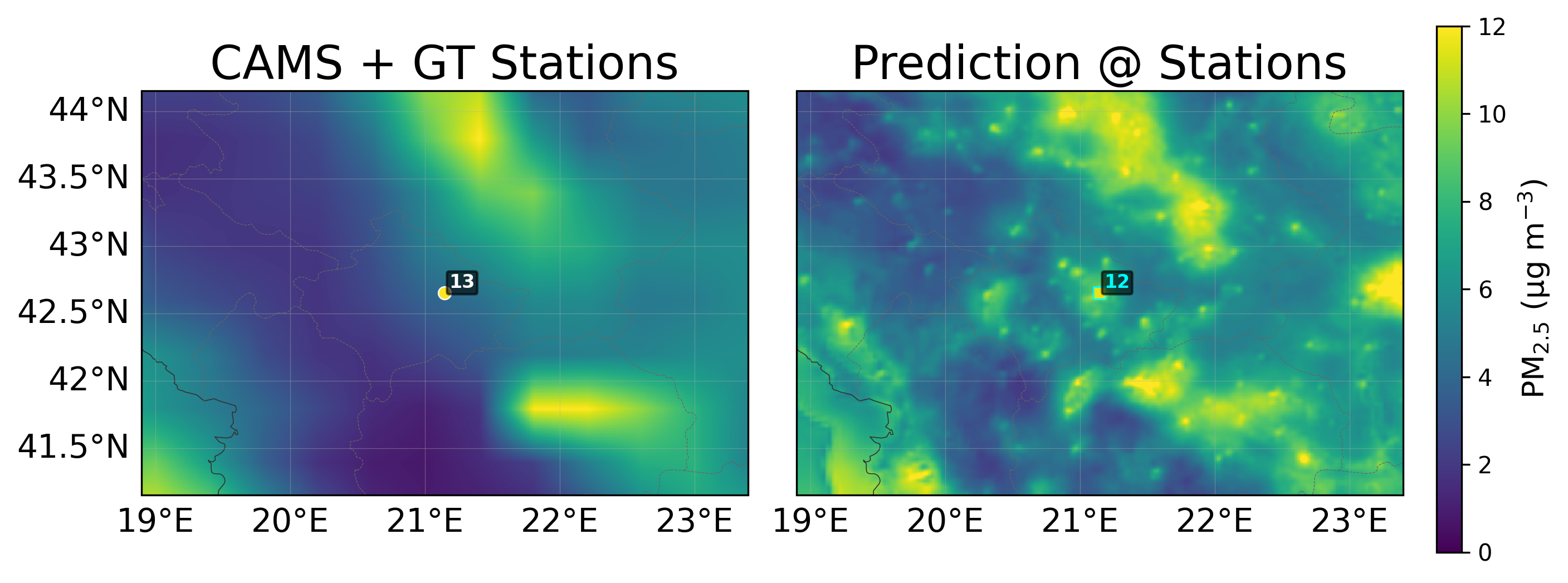}
    \end{tabular}
    \caption{Regional PM$_{2.5}$ predictions over Europe.}
    \label{fig:pm25_regions}
\end{figure*}

Figures~\ref{fig:eu_pm25_20200805_20} and~\ref{fig:pm25_regions} show the qualitative results of our high-resolution PM${2.5}$ estimation. Compared with the original CAMS Forecast field at a spatial resolution of 0.4$^\circ$, our model produces predictions on a 0.01$^\circ$ grid, revealing much finer local spatial structures. In addition to increasing the spatial granularity, the model also uses station observations to locally correct the CAMS baseline, leading to more realistic concentration patterns around observed regions. Figure~\ref{fig:eu_pm25_zoomin} further demonstrates that the proposed method can generate visually consistent high-resolution PM${2.5}$ maps over large-scale geographical regions.

\begin{figure}[t]
    \centering
    \includegraphics[width=\linewidth]{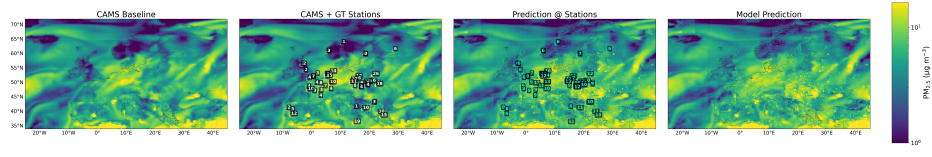}
    \caption{PM$_{2.5}$ distribution over Europe.}
    \label{fig:eu_pm25_20200805_20}
\end{figure}

\begin{figure}[!htbp]
    \centering

    \begin{subfigure}[t]{0.49\textwidth}
        \centering
        \includegraphics[width=\linewidth]{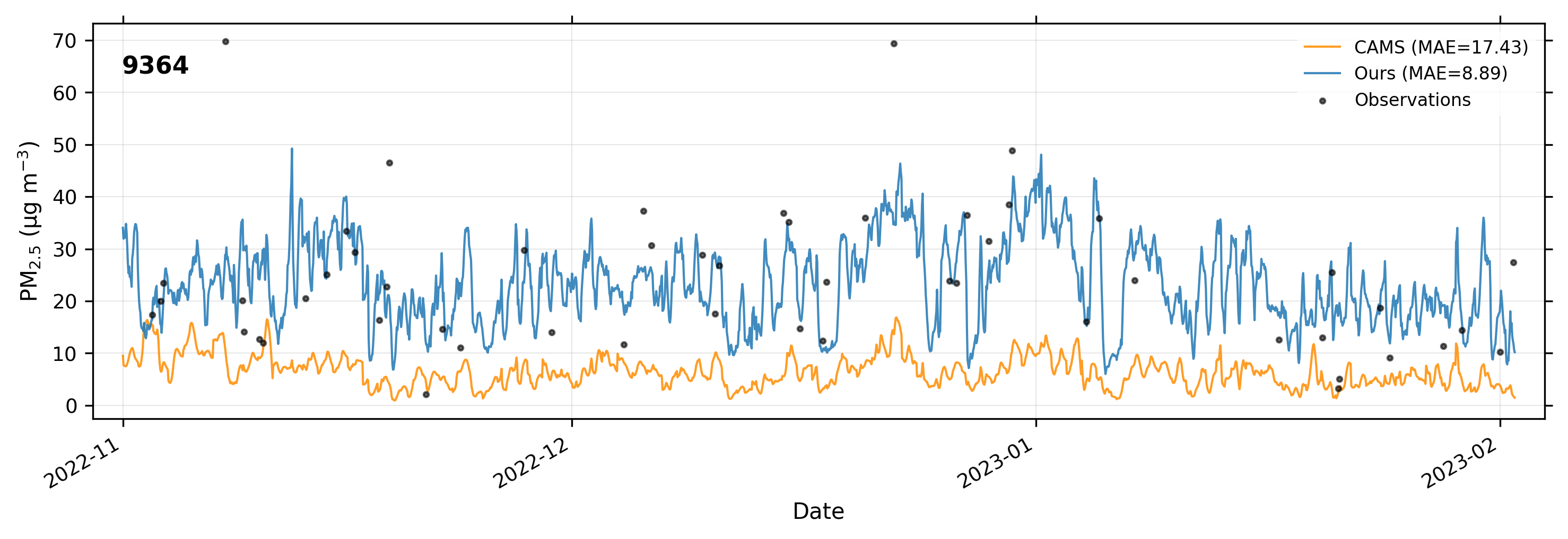}
        \caption{High-concentration case.}
        \label{fig:timeseries_high}
    \end{subfigure}
    \hfill
    \begin{subfigure}[t]{0.49\textwidth}
        \centering
        \includegraphics[width=\linewidth]{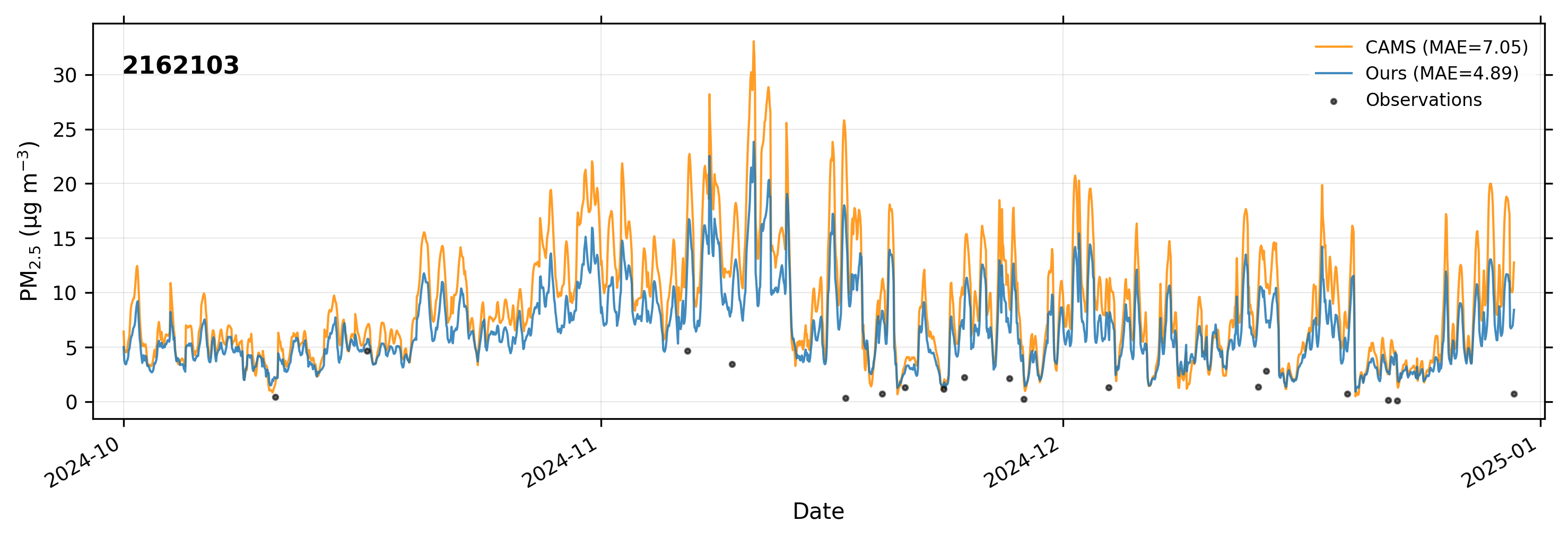}
        \caption{Low-concentration case.}
        \label{fig:timeseries_low}
    \end{subfigure}

    \caption{
    Temporal comparison between ground-truth PM$_{2.5}$ observations, CAMS, and STARQ at two representative validation stations.
    }
    \label{fig:timeseries_case_studies}
\end{figure}

To assess whether our model can track temporal variations at fixed sites, we present time-series case studies at two representative validation stations, shown in Figure~\ref{fig:timeseries_case_studies}. 
This comparison includes the CAMS baseline, the proposed model prediction, and the station ground-truth observations at the same locations. 
Figure~\ref{fig:timeseries_case_studies}(a) presents the results for a station with pronounced pollution episodes. In this case, CAMS systematically underestimates high-PM$_{2.5}$ values, especially during winter peaks. 
In contrast, the proposed model better follows the temporal variability of the ground-truth observations and reduces the MAE from 17.43 to 8.89. 
Figure~\ref{fig:timeseries_case_studies}(b) focuses on a station with relatively low concentration levels. Here, CAMS tends to overestimate PM$_{2.5}$ over the time period. 
The proposed model corrects this positive bias and produces predictions closer to the observations, reducing the MAE from 7.05 to 4.89.

These results suggest that CAMS errors are location-dependent: the coarse-resolution baseline can underestimate polluted episodes at some locations while overestimating concentrations at others. 
By learning spatial and temporal side information from other monitoring stations and side information, the proposed model mitigates these site-specific biases and improves prediction accuracy at unseen timestamps and locations.

\subsection{Channel importance}
We assess local channel importance on a representative validation region using three complementary methods: (i) leave-one-channel-out ablation, monitoring the increase in MAE after removing a channel, (ii) gradient-based sensitivity, measuring model output sensitivity to input channel perturbations, and (iii) permutation-based importance, evaluating how much the model relies on spatially and temporally aligned information in each channel. Each method is described in Section D of the Supplementary Material. 

Figure~\ref{fig:local_channel_importance} summarizes the three local channel-importance analyses. 
The ablation and permutation results both measure changes in MAE and therefore directly reflect whether a channel improves predictive performance in this local region. 
Both methods identify CAMS PM$_{2.5}$ and GHSL built-up features as the dominant inputs. 
This is consistent with the role of CAMS as a coarse atmospheric prior and GHSL built-up variables as proxies for urban morphology and emission-related spatial structure. 
Other auxiliary variables, such as CLC land cover, ERA5 wind components, and DEM elevation, provide smaller but generally positive contributions. 
In contrast, MODIS AOD and GHSL population show slightly negative importance in this local test, suggesting that these channels may be noisy, redundant, or less aligned with the local station-level PM$_{2.5}$ correction.

The gradient-based sensitivity assigns non-negligible sensitivity to most channels, indicating that the model responds to multiple inputs in this region. However, this metric should be interpreted differently from the MAE-based scores: a high gradient sensitivity reflects local responsiveness to a channel, but does not necessarily mean that the channel improves predictive accuracy. A channel can strongly influence the model's output while still contributing noise or redundant information, which would not be captured by gradient-based measures alone. Therefore, the three methods are best viewed as complementary: ablation and permutation importance reveal which channels are beneficial for performance, while gradient sensitivity highlights which channels the model relies on most, regardless of their effect on error.

\begin{figure*}[!t]
    \centering

    \begin{subfigure}[t]{0.32\textwidth}
        \centering
        \includegraphics[width=\linewidth]{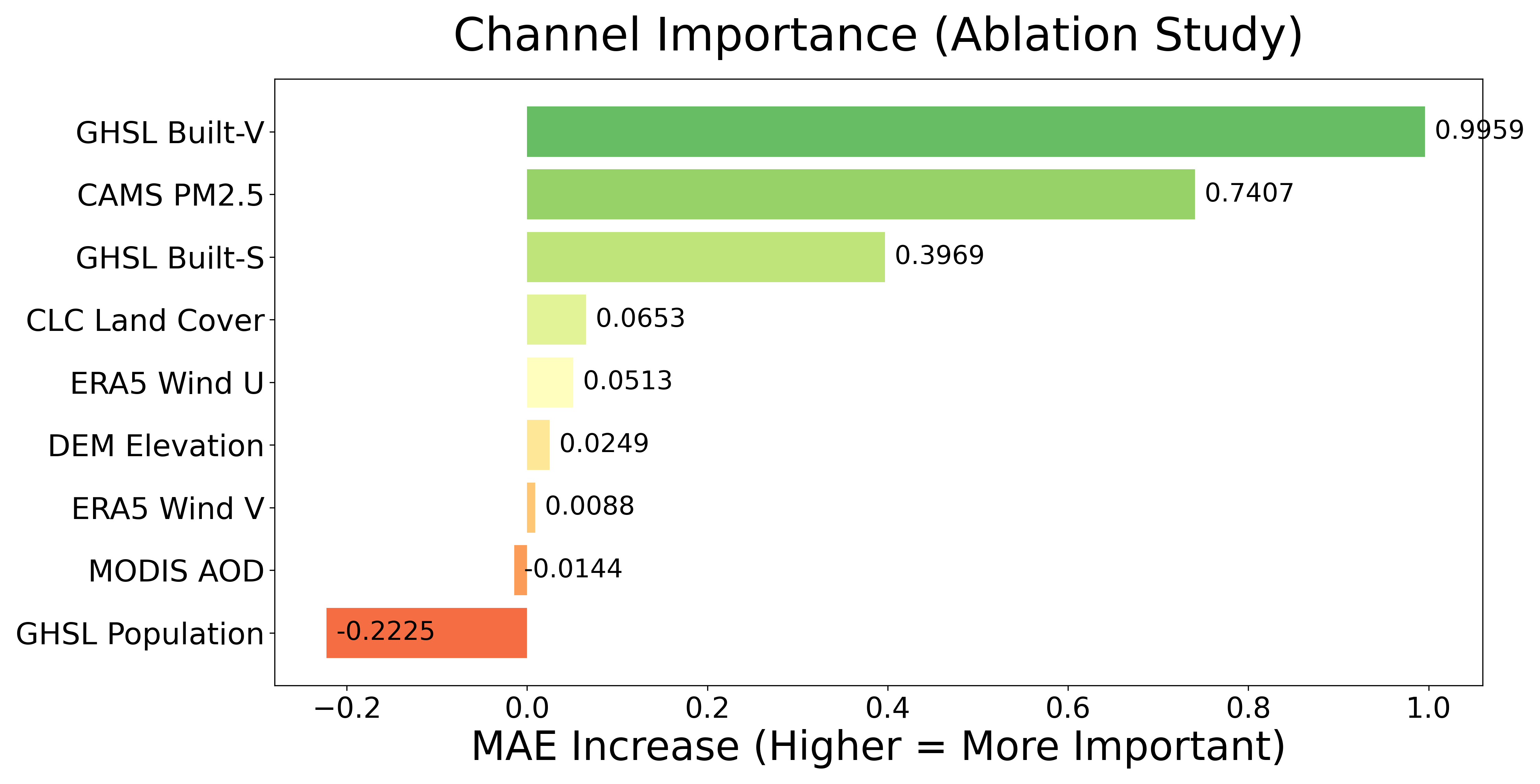}
        \caption{Ablation.}
        \label{fig:channel_importance_ablation}
    \end{subfigure}
    \hfill
    \begin{subfigure}[t]{0.32\textwidth}
        \centering
        \includegraphics[width=\linewidth]{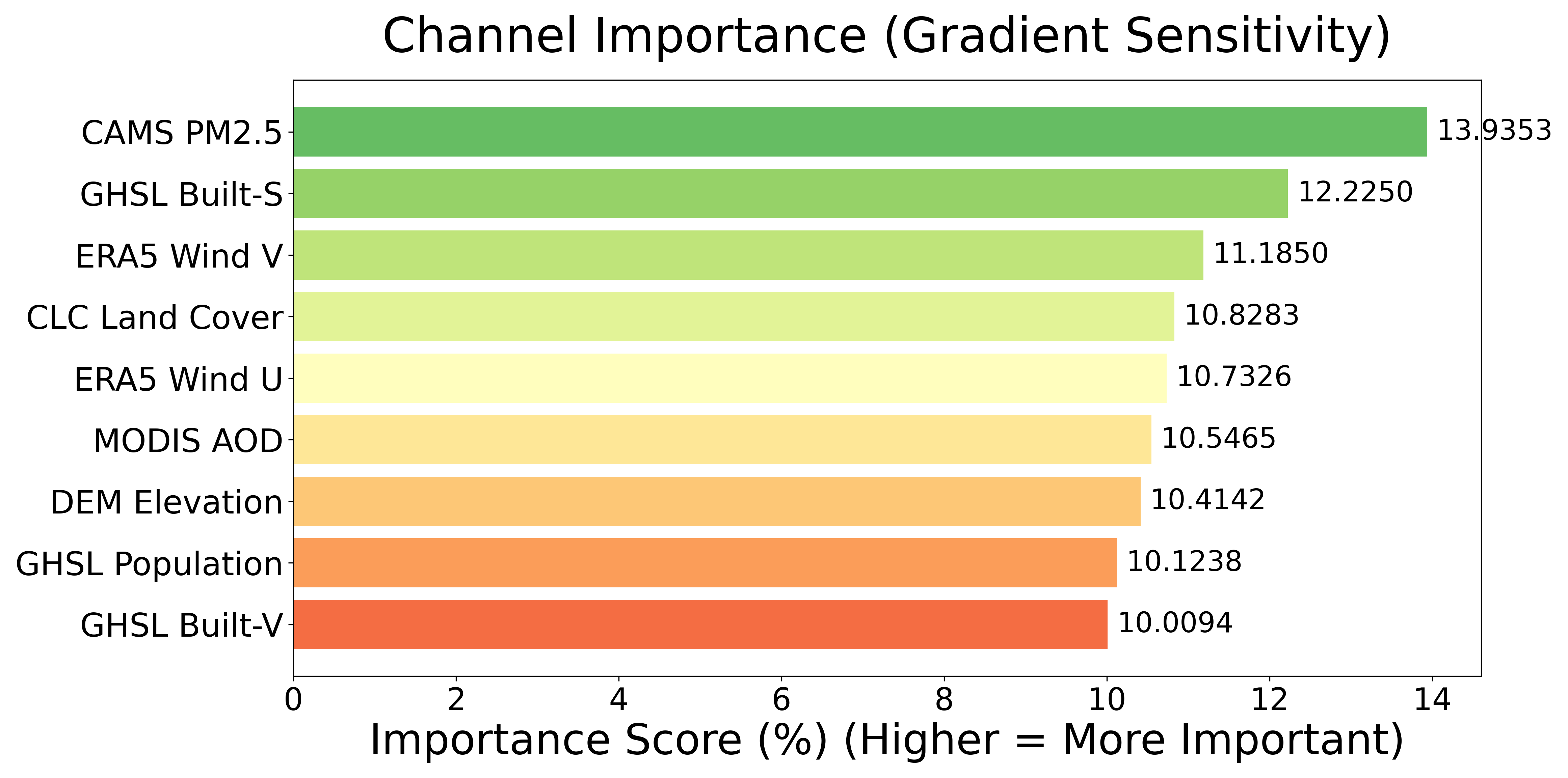}
        \caption{Gradient sensitivity.}
        \label{fig:channel_importance_gradient}
    \end{subfigure}
    \hfill
    \begin{subfigure}[t]{0.32\textwidth}
        \centering
        \includegraphics[width=\linewidth]{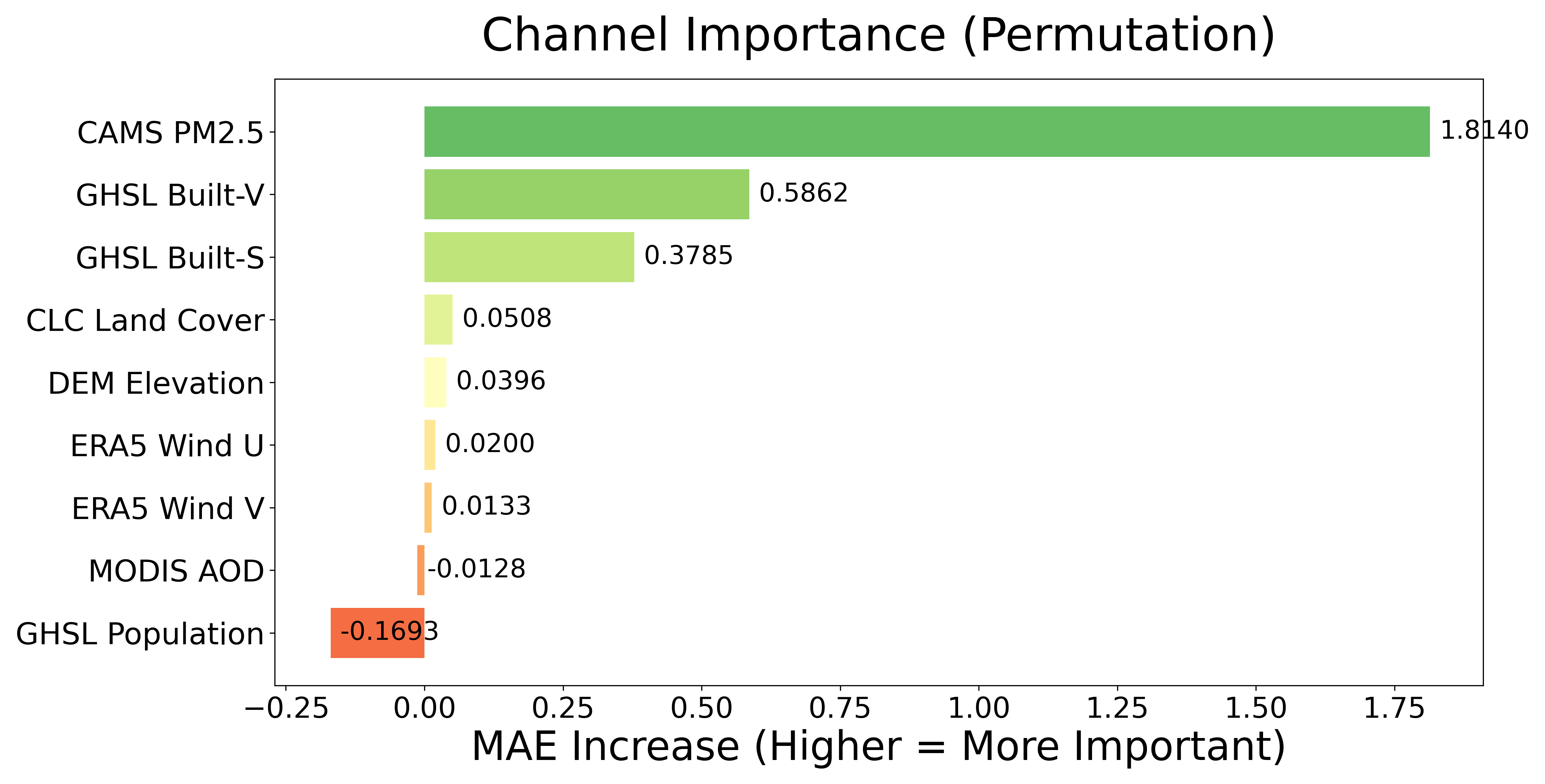}
        \caption{Permutation.}
        \label{fig:channel_importance_permutation}
    \end{subfigure}

    \caption{
    Local channel-importance analysis on a representative validation region.
    }
    \label{fig:local_channel_importance}
\end{figure*}

\section{Conclusion}
We introduce STARQ, a station-guided downscaling framework for bias-corrected PM$_{2.5}$ estimation over Europe. The proposed method combines the spatial completeness of CAMS with the point-level accuracy of OpenAQ observations, while incorporating auxiliary variables that describe human activity, land cover, topography, aerosol loading, and meteorological transport. By using Gaussian-kernel interpolation and pseudo-label propagation, the framework converts sparse station measurements into dense supervision and enables time-agnostic spatial correction at arbitrary timestamps.

\looseness-1Experiments on the European test set demonstrate that the proposed model consistently improves upon the CAMS baseline in station-level evaluation, achieving lower MAE and RMSE and a substantially higher $R^2$. Qualitative results further show that the model generates spatially detailed PM$_{2.5}$ fields at 0.01$^\circ$ resolution and produces smoother, more realistic local structures than the original coarse CAMS field. The time-series case studies also suggest that the model can mitigate location-dependent CAMS biases, including both underestimation during high-pollution episodes and overestimation in low-concentration regions. Additional ablation and channel-importance analyses indicate that CAMS PM$_{2.5}$ and built-up features provide particularly important information, while other environmental and meteorological variables contribute complementary spatial context.

Several limitations remain. The quality of supervision depends on the coverage and reliability of OpenAQ station measurements, which are spatially uneven and may contain outliers. In addition, the current pseudo-label design relies on fixed kernel-based propagation, which may not fully capture complex pollutant transport under varying meteorological conditions. Future work will explore adaptive and uncertainty-aware kernels, stronger quality-control strategies for station observations, and broader evaluation across additional pollutants, regions, and extreme pollution events.

\section*{Acknowledgements}

GW, SF, and SZ were supported by the EPSRC Turing AI Fellowship (Grant Ref: EP/Z534699/1): Generative Machine Learning Models for Data of Arbitrary Underlying Geometry (MAGAL).
AD and MN were supported through the TensorICE project (EXCELLENCE/\allowbreak0524/\allowbreak0407), which is implemented under the social cohesion programme "THALIA 2021-2027", co-funded by the European Union through the Research and Innovation Foundation of Cyprus.

\clearpage
%
%
\bibliographystyle{splncs04}
\bibliography{main}
\clearpage

\appendix
\section*{Supplementary Material}
\section{Additional Data Processing Details}
\subsection{Side Information Data Sources} 
    \label{ap:data_source}

The data used to train the model is obtained from various sources, including side information and ground-truth station measurements.
A diverse set of geospatial and atmospheric datasets was employed to analyse fine particulate air pollution (PM$_{2.5}$) and its driving factors. These datasets span atmospheric composition forecasts, human settlement and population data, land cover information, topography, satellite-derived aerosol measurements, and meteorological reanalysis.

\subsubsection{CAMS Global Atmospheric Composition Forecasts.} 
The Copernicus Atmosphere Monitoring Service (CAMS) Global Atmospheric Composition Forecasts provide predictions of atmospheric pollutant concentrations worldwide. The CAMS forecasts are generated using an atmospheric model based on physics and chemistry, which assimilates satellite observations to ensure accuracy. CAMS issues forecasts twice daily, each providing hourly predictions of more than 50 chemical species and seven aerosol types. Notably, the CAMS forecasts include particulate matter outputs such as surface-level PM$_{2.5}$ (particulate matter with a diameter smaller than 2.5~µm), expressed in units of mass concentration. The horizontal resolution of the global forecast is approximately 0.4$^\circ\times$0.4$^\circ$ ($\approx$40$\times$40~km). We utilised the CAMS PM$_{2.5}$ forecast fields as the air quality input, providing an initial estimate of pollutant concentrations across our study domain. This ensured that our PM$_{2.5}$ analysis had a consistent global prior informed by both modelling and satellite observation integration, which can be combined with station observations.

\subsubsection{Global Human Settlement Layer (GHSL).}

The Global Human Settlement Layer is a collection of spatial datasets supported by the Joint Research Centre (JRC) and the Directorate-General for Regional and Urban Policy (DG REGIO) of the European Commission, together with the international partnership GEO Human Planet Initiative. GHSL produces global spatial information, evidence-based analytics, and knowledge describing the human presence on the planet. These data are useful for incorporating anthropogenic factors into environmental analyses, as they quantify the built environment and population distribution globally. In particular, we employed three GHSL products to capture human-related variables relevant to air pollution:

\begin{itemize}
    \item \textbf{GHS-BUILT-S (Built-up Surface)}  
    GHS-BUILT-S depicts the distribution of built-up surfaces, expressed as the number of square metres. It describes the extent of built-up areas, providing the proportion of each cell that is covered by built-up structures. This built-up surface layer helps identify urbanised areas and infrastructure, which are often associated with higher emissions and concentrations of PM$_{2.5}$.

    \item \textbf{GHS-BUILT-V (Built-up Volume)}  
    GHS-BUILT-V provides estimates of the distribution of built-up volume (in cubic metres) within each grid cell. It combines information on built-up area and building height to quantify the three-dimensional volume of urban structures. This allows us to account not only for the horizontal footprint of urban areas but also for their vertical extent (e.g., differences between high-rise and low-rise buildings), which may influence local pollution emissions and dispersion.

    \item \textbf{GHS-POP (Population Grid)}  
    The GHSL population grid depicts the distribution of human population as the number of people per cell. This dataset enables analysis of the spatial relationship between population distribution and pollution sources, as well as the diffusion of pollution influenced by human presence. By incorporating population density, our assessment can identify areas where high PM$_{2.5}$ concentrations coincide with large populations.
\end{itemize}

Together, these GHSL layers enable a comprehensive characterisation of human activities. In our study, they are used to incorporate anthropogenic factors into the PM$_{2.5}$ analysis.

\subsubsection{CORINE Land Cover.}

CORINE Land Cover (CLC) is a pan-European land cover inventory that provides land use and land cover information. CORINE provides a harmonised classification of land cover across Europe with 44 thematic classes, ranging from broad forested areas to individual vineyards.
Land cover is an important factor in air quality studies, as it influences emission sources and affects pollutant dispersion---for instance, depending on whether an area is industrial, agricultural, or forested. By using CLC, we can distinguish urban areas from natural land. The inclusion of CLC therefore provides contextual information about the environment in which air pollution occurs, enhancing the robustness of our learning of PM$_{2.5}$ patterns.

\subsubsection{EU-DEM (Digital Elevation Model).}

EU-DEM is a high-resolution digital elevation model covering Europe, representing ground elevations and topography. We included elevation data because terrain influences atmospheric flow and pollution distribution. For example, valleys can trap pollutants, leading to higher PM$_{2.5}$ concentrations, while elevated or exposed areas may experience greater dispersion. By integrating EU-DEM, our analysis captures the effects of elevation by examining whether high pollution levels coincide with specific topographic features and by including elevation as a predictor in statistical models of PM$_{2.5}$. 

\subsubsection{MODIS MCD19A2 Aerosol Optical Depth.}
To incorporate satellite observations of atmospheric aerosols, we used the MODIS MCD19A2 product. The MCD19A2 Version 6.1 dataset is a Moderate Resolution Imaging Spectroradiometer (MODIS) Terra and Aqua combined Multi-Angle Implementation of Atmospheric Correction (MAIAC) Land Aerosol Optical Depth (AOD) gridded Level~2 product, produced daily at a 1~km pixel resolution. Specifically, MCD19A2 provides aerosol optical thickness measurements for the blue band (AOD at 0.47~µm) and the green band (AOD at 0.55~µm) for each day. By using this dataset, we incorporated an independent observational perspective on particulate pollution since PM$_{2.5}$ constitutes a major component of atmospheric aerosols and is therefore correlated with AOD. This is particularly useful in regions or during periods where ground-based measurements are limited. Its daily resolution also captures day-to-day variability in aerosol levels, which is relevant for short-term air quality assessment.

\subsubsection{ERA5 Reanalysis (Wind Data).}

Meteorological conditions play a crucial role in air pollution dispersion. Therefore, we incorporated wind data from ERA5, the fifth-generation global atmospheric reanalysis produced by the European Centre for Medium-Range Weather Forecasts (ECMWF). ERA5 provides hourly estimates of a wide range of atmospheric variables at a spatial resolution of approximately 0.25$^\circ$ on a global scale. Specifically, we extracted near-surface wind fields (10~m wind components) from the ERA5 single-level dataset. The eastward (\textit{u10}) and northward (\textit{v10}) 10~m wind components describe the ventilation and transport of pollutants: stronger winds generally enhance dispersion, whereas stagnant conditions promote pollutant accumulation. The ERA5 fields provide reliable meteorological inputs corresponding to the temporal and spatial coverage of our PM$_{2.5}$ data.
In our study, we used ERA5 wind speed and direction to investigate how local and regional airflow patterns influence the spatiotemporal variability of PM$_{2.5}$. Including ERA5 meteorological data thus provides a dynamic physical context for interpreting PM$_{2.5}$ patterns, helping to distinguish whether elevated pollutant concentrations result from emission sources or from meteorological stagnation.

\subsection{Coordinate and Resolution Conversions.}
These datasets are provided by different organisations and therefore use different coordinate systems and spatial resolutions. In this study, we adopt a fine resolution of 0.01$^\circ$.
Consequently, we need a way to reconcile the differences among datasets. A practical solution is to reproject all the side information to our desired coordinate system and resolution beforehand and then load the standardised data during training.

For the station data, OpenAQ provides point measurements that are tied to specific latitude–longitude locations. We also need to project each point onto a 2-D image representation (using the same coordinate system and resolution as the side information) to enable patch-to-patch training.
For each patch, we must include the measurement values from all stations falling within that patch, along with the corresponding mask locations.

However, including 10 years of data from stations, satellites, and other anthropogenic data sources across Europe is a large-scale data-processing challenge, and speeding up the data-generation process is essential. We therefore distribute the work across multiple CPU cores, assigning each worker a different hour within the same European region so that the workflow runs in parallel.

\section{Additional Training Details}
\subsection{Input Construction and Label Projection}
\label{sec:training_strategy}
For each timestamp, we construct the complete model input by concatenating
the hourly atmospheric variables with the slowly varying geospatial
variables, which are typically updated annually. OpenAQ PM$_{2.5}$
observations are projected onto the common $0.01^\circ$ grid and matched to
the corresponding hourly input timestamp; since the side information is at
hourly resolution, minute-level differences in OpenAQ timestamps are
discarded and only the hour is retained. In the rare case where multiple
observations fall into the same grid cell at the same timestamp, their
values are averaged and a single record is retained. This has negligible
impact in practice, as such stations are usually within about 1~km of each
other and report very similar pollution values. The station-level
train/validation/test split is then applied to determine the observations
available for supervision at each timestamp.

\subsection{Patch-Based Sampling}
Processing the full European domain directly is computationally prohibitive,
as loading the entire map into GPU memory is infeasible. We therefore adopt
patch-based training with $64\times64$ input patches; at the $0.01^\circ$
($\approx$1~km) target resolution, each patch covers an approximate spatial
extent of $64\,\mathrm{km}\times64\,\mathrm{km}$. For each timestamp, a
station is selected as the patch centre and a $64\times64$ window is cropped
around it, with random horizontal and vertical offsets within a controlled
range to increase input diversity. All stations located inside the window
are used as supervision signals for that patch.

Because the stations available at a given timestamp may span a wide
geographical area across Europe, we sample multiple station-centred patches
from the same timestamp within each epoch, allowing the model to learn from
all station locations. Repeatedly sampling the same station is also
beneficial, since the random spatial offset introduces local variation and
improves robustness. To preserve the large-scale spatial structure of the
CAMS prior, we additionally sample background patches without station
supervision. For each patch, the corresponding pseudo-label is constructed
using different Gaussian kernel designs.

Each sample is recorded together with its timestamp and, for station
patches, the station coordinates, in the form
\[
(\text{timestamp\_id}, \text{station}, \text{station coordinates})
\]
or
\[
(\text{timestamp\_id}, \text{background}).
\]
This ensures that every station can be sampled at least once as the centre
of a training patch, so that all available stations are traversed.

\subsection{Bucketed Timestamp Sampling}
Our batch size is large and training is conducted on multiple GPUs, so many
timestamp files may be opened simultaneously. Each file contains nine
channels over the full European domain and exceeds 400~MB. Although the
files are accessed through file handles rather than being fully loaded into
memory, opening many large files at once causes heavy I/O pressure, leaving
the GPUs under-utilised while they wait for data loading. In the worst case,
each training step may need to access approximately 
\[
\text{batch size} \times \text{number of GPUs}
\]
different files, which significantly slows down training.

Loading all patches from a single timestamp minimises the number of files
opened and improves I/O locality, but it sacrifices temporal diversity:
PM$_{2.5}$ is highly time-sensitive and the side information can vary
substantially across timestamps. A batch drawn from a single timestamp may
cause the model to overfit to a specific time point and may produce gradients
that vary significantly between steps, leading to unstable training.
Conversely, fully random sampling provides high data diversity but causes
extremely slow I/O. Our goal is to balance these two extremes.

Inspired by bucket sampling in NLP, where samples of similar length are
grouped together to reduce padding, we group patches from a small number of
timestamps into the same bucket. Within a bucket, several patches reuse the
same timestamp files, so the data only needs to be read from storage once,
while later accesses benefit from the file cache. The complete procedure is
summarised in Algorithm~\ref{alg:bucket_sampler}.

\begin{algorithm}[t]
\caption{Bucketed Timestamp Sampling}
\label{alg:bucket_sampler}
\begin{algorithmic}[1]
\Require Timestamp set $\mathcal{T}$, sample sets
$\{\mathcal{S}_t\}_{t\in\mathcal{T}}$, bucket size $K$
\Ensure Ordered mini-batches for one training epoch

\State Randomly shuffle the timestamps in $\mathcal{T}$
\State Partition the shuffled timestamps into buckets
       $\{\mathcal{B}_1,\ldots,\mathcal{B}_M\}$,
       where $|\mathcal{B}_m|\leq K$

\For{$m=1,\ldots,M$}
    \State Collect all samples associated with the timestamps in the bucket:
    \[
        \mathcal{S}_m
        \leftarrow
        \bigcup_{t\in\mathcal{B}_m}\mathcal{S}_t
    \]
    \State Randomly shuffle the samples in $\mathcal{S}_m$
    \State Split $\mathcal{S}_m$ into mini-batches
    \State Yield the mini-batches from $\mathcal{S}_m$ sequentially
\EndFor
\end{algorithmic}
\end{algorithm}

At the beginning of every epoch, the timestamp order is reshuffled, so that
different timestamps are grouped together across epochs. Samples are also
shuffled within each bucket, preventing consecutive mini-batches from being
dominated by a single station or location. Processing each bucket
sequentially restricts data access to a small set of timestamp files,
thereby improving I/O locality while retaining temporal variation within
each bucket.

The bucket size, defined as the number of timestamps in each bucket,
controls this trade-off. A size of one maximises I/O efficiency but
degenerates into single-timestamp sampling and loses diversity, whereas a
very large size approaches fully random sampling. We use a bucket size of
$K=4$ in all experiments. Preliminary training runs revealed that this choice reduces the training time from
around 10 days per epoch under fully shuffled sampling to approximately
1--2 days per epoch.

\subsubsection{Worked Example: DataLoader Construction.}

The final data loader follows a bucketed timestamp sampling procedure. Consider a simple example with 12 timestamps and a bucket size $K=4$.
First, all timestamps are placed into a list and randomly shuffled:
\[
[\text{ts}_0, \text{ts}_1, \ldots, \text{ts}_{11}]
\]
may become
\[
[\text{ts}_7, \text{ts}_2, \text{ts}_{11}, \text{ts}_5,
\text{ts}_0, \text{ts}_9, \text{ts}_3, \text{ts}_8,
\text{ts}_1, \text{ts}_{10}, \text{ts}_6, \text{ts}_4].
\]
The shuffled timestamp list is then divided into consecutive chunks of size $K=4$:
\[
\text{Group A} = [\text{ts}_7, \text{ts}_2, \text{ts}_{11}, \text{ts}_5],
\]
\[
\text{Group B} = [\text{ts}_0, \text{ts}_9, \text{ts}_3, \text{ts}_8],
\]
\[
\text{Group C} = [\text{ts}_1, \text{ts}_{10}, \text{ts}_6, \text{ts}_4].
\]
Next, all samples from the timestamps within the same bucket are collected. For example, suppose each timestamp contains around 500 samples, including station patches and background patches. Group A may contain
\[
520 + 480 + 510 + 490 = 2000
\]
samples in total. These 2000 samples are then randomly shuffled within the group.
Finally, all shuffled groups are concatenated sequentially:
\[
[\text{shuffled Group A} \mid \text{shuffled Group B} \mid \text{shuffled Group C}].
\]
The PyTorch DataLoader then splits this final index sequence into mini-batches, for example with a batch size of 384.

In short, the procedure first shuffles the timestamp order, then groups every $K$ timestamps into one bucket, and finally shuffles all patches within each bucket. The timestamp grouping is different in every epoch, which avoids a fixed sampling pattern. Within each bucket, patch-level shuffling ensures that a batch is not dominated by samples from a single station. Processing buckets sequentially improves I/O locality because consecutive batches tend to access the same small set of timestamp files.

\section{Additional Evaluation Details}
\subsection{Evaluation Metrics}

The predicted values at station locations are compared against the corresponding ground-truth observations. The model is optimised using a root mean squared error (RMSE) loss, and the
checkpoint with the lowest mean absolute error (MAE) on the spatiotemporally held-out OpenAQ validation set is selected as the best
model.
Based on the matched prediction--observation pairs, we also evaluate model performance using the coefficient of determination ($R^2$).

For MAE and RMSE, lower values indicate better performance, as they measure the magnitude of prediction errors. 
MAE is more robust to outliers, while RMSE penalises larger errors more heavily due to the squared term.
The coefficient of determination $R^2$ ranges from $-\infty$ to 1. 
A value of $R^2 = 1$ indicates perfect prediction, while $R^2 = 0$ means the model performs no better than predicting the mean of the ground-truth values. 
Negative values indicate that the model performs worse than this baseline.
Overall, a good model is expected to achieve low MAE and RMSE and high $R^2$.

\subsection{Visualisation Procedure}
For the visualisations, we perform sliding-window inference with window size $W_s$ and overlap $O$, using stride $s=W_s-O$.
Let the input raster be $x\in\mathbb{R}^{C\times H\times W}$ and the model be $f_\theta$, producing a local prediction for each window.
For each window $k$ with top-left corner $(h_k,w_k)$, we extract a tile
\[
x_k = x[:,\, h_k:h_k+a_k,\; w_k:w_k+b_k],
\]
(where $a_k\times b_k$ is the valid window size; boundary windows may have $a_k<W_s$ and/or $b_k<W_s$ and are reflect-padded to $W_s\times W_s$ before inference).
The model outputs a local prediction $\hat{y}_k = f_\theta(x_k)$ on the valid region $\Omega_k=\{0,\dots,a_k-1\}\times\{0,\dots,b_k-1\}$.

To suppress boundary artefacts, we apply a distance-based fading weight.
Let $(u,v)\in\Omega_k$ be local coordinates and define the minimum distance to window edges
\[
d(u,v)=\min\!\bigl(u,\; v,\; a_k-1-u,\; b_k-1-v\bigr).
\]
With fading width $F=O/2$, we use a quadratic fade
\[
\omega_k(u,v)=
\begin{cases}
\left(\dfrac{d(u,v)}{F}\right)^2, & d(u,v)<F,\\
1, & \text{otherwise}.
\end{cases}
\]
Overlapping predictions are merged by normalised weighted averaging:
\[
\hat{y}(h,w)=
\frac{
\sum\limits_{k:(h,w)\in\mathcal{R}_k}\omega_k(h-h_k,w-w_k)\,\hat{y}_k(h-h_k,w-w_k)
}{
\sum\limits_{k:(h,w)\in\mathcal{R}_k}\omega_k(h-h_k,w-w_k)
},
\]
where $\mathcal{R}_k=\{(h,w): (h-h_k,w-w_k)\in\Omega_k\}$ is the footprint of window $k$ in the global raster.
This design ensures smooth transitions between neighbouring patches and avoids visible discontinuities at patch boundaries.

\section{Channel Importance in the Italian Region}
For an input patch $\mathbf{X} \in \mathbb{R}^{C \times H \times W}$, where the $C$ channels include CAMS PM$_{2.5}$ and auxiliary geospatial and meteorological variables, a trained CNN predicts $\hat{\mathbf{Y}} = f_{\theta}(\mathbf{X})$. Since station observations are sparse, errors are evaluated only at observed pixels using the valid mask $\mathbf{M} \in \{0,1\}^{H \times W}$. The masked MAE is $\mathcal{E}(\hat{\mathbf{Y}}, \mathbf{Y}; \mathbf{M}) = \frac{\sum_{p \in \Omega} M_p |\hat{Y}_p - Y_p|}{\sum_{p \in \Omega} M_p}$, and the full-channel baseline error is $\mathcal{E}_0 = \mathcal{E}(f_{\theta}(\mathbf{X}), \mathbf{Y}; \mathbf{M})$.
We compute channel importance using three complementary methods.

\subsection{Leave-One-Channel-Out Ablation}
For each channel $c$, we construct an ablated input $\mathbf{X}^{(-c)}$ by setting the $c$-th channel to zero while keeping all other channels unchanged:
\[
    X^{(-c)}_j =
    \begin{cases}
        0, & j = c, \\
        X_j, & j \neq c.
    \end{cases}
\]
The ablation importance score is defined as the increase in MAE after removing channel $c$:
\[
    I^{\mathrm{abl}}_c
    =
    \mathcal{E}(f_{\theta}(\mathbf{X}^{(-c)}), \mathbf{Y}; \mathbf{M})
    -
    \mathcal{E}_0.
\]
A positive value indicates that removing this channel degrades the model performance, suggesting that the channel provides useful information. A negative value means that removing the channel slightly improves the local MAE, which may indicate noise, redundancy, or local overfitting with the station-level correction.

\subsection{Gradient-Based Sensitivity}
The second method measures the local sensitivity of the loss with respect to each input channel. We use the masked MSE loss
\[
    \mathcal{L}
    =
    \frac{
        \sum_{p \in \Omega} M_p
        \left(f_{\theta}(\mathbf{X})_p - Y_p\right)^2
    }{
        \sum_{p \in \Omega} M_p
    }.
\]
For each channel $c$, we compute the mean absolute input gradient:
\[
    G_c
    =
    \frac{1}{|\mathcal{D}|}
    \sum_{\mathbf{X} \in \mathcal{D}}
    \frac{1}{HW}
    \sum_{p \in \Omega}
    \left|
        \frac{\partial \mathcal{L}}{\partial X_{c,p}}
    \right|,
\]
where $\mathcal{D}$ denotes the sampled validation patches. The gradient importance is then normalised across all channels:
\[
    I^{\mathrm{grad}}_c
    =
    100 \times
    \frac{G_c}{\sum_{j=1}^{C} G_j}.
\]
Unlike ablation and permutation, this score does not directly measure the effect on MAE. Instead, it measures how sensitive the model output is to local perturbations in each input channel.

\subsection{Permutation-Based Importance}
The third method evaluates how much the model relies on the spatially and temporally aligned information in each channel. For channel $c$, we randomly permute this channel across validation samples while leaving the remaining channels unchanged:
\[
    X^{\mathrm{perm}(c,r)}_{b,j}
    =
    \begin{cases}
        X_{\pi_r(b),c}, & j = c, \\
        X_{b,j}, & j \neq c,
    \end{cases}
\]
where $b$ indexes validation samples, $\pi_r$ is a random permutation for repeat $r$, and $r=1,\dots,R$. The permutation importance score is the average MAE increase over $R$ repeats:
\[
    I^{\mathrm{perm}}_c
    =
    \frac{1}{R}
    \sum_{r=1}^{R}
    \left[
    \mathcal{E}
    \left(
        f_{\theta}(\mathbf{X}^{\mathrm{perm}(c,r)}),
        \mathbf{Y};
        \mathbf{M}
    \right)
    -
    \mathcal{E}_0
    \right].
\]
The corresponding standard deviation across repeats is also recorded to quantify the variability introduced by random permutation.

\end{document}